%% file: main.tex
\documentclass[conference]{IEEEtran}

\usepackage{amsmath}
\usepackage{amssymb}
\usepackage{epsfig}
\usepackage{times}

\usepackage[noadjust]{cite}

\usepackage[colorlinks, bookmarks=false, breaklinks=true]{hyperref}
\hypersetup{
    colorlinks,
    linkcolor={black},
    citecolor={black},
    urlcolor={black}
}

\usepackage[main=english, ngerman]{babel}
\usepackage[T1]{fontenc}
\usepackage[utf8]{inputenc}
\usepackage{mathtools, amsfonts}
\usepackage[style=english]{csquotes}

\usepackage[dvipsnames]{xcolor}
\usepackage{tikz}
\usetikzlibrary{arrows, calc, external, fit, positioning, shapes, spy}

\usepackage{pgfplots}
\pgfplotsset{compat=newest}
\usepgfplotslibrary{colorbrewer}

\usepackage{
	array,
	multirow,
	subfig,
	xfrac
}

\usepackage[export]{adjustbox}

\usepackage{graphicx, graphbox}
\graphicspath{{figs/}, {title/}}

\newcolumntype{P}[1]{>{ \centering  \arraybackslash }p{#1}}
\newcolumntype{Q}[1]{>{ \raggedleft \arraybackslash }p{#1}}
\newcolumntype{M}[1]{>{ \centering  \arraybackslash }m{#1}}
\newcolumntype{N}[1]{>{ \raggedleft \arraybackslash }m{#1}}
\newcolumntype{B}[1]{>{ \centering  \arraybackslash }b{#1}}
\newcolumntype{C}[1]{>{ \raggedleft \arraybackslash }b{#1}}

\usepackage{fancyhdr}
\fancypagestyle{psp1}{
	
	\fancyfoot[C]{IECON 2020 - \copyright~2020 IEEE}
}

\usepackage{scalerel}
\usetikzlibrary{svg.path}

\definecolor{orcidlogocol}{HTML}{A6CE39}
\tikzset{
    orcidlogo/.pic={
        \fill[orcidlogocol] svg{M256,128c0,70.7-57.3,128-128,128C57.3,256,0,198.7,0,128C0,57.3,57.3,0,128,0C198.7,0,256,57.3,256,128z};
        \fill[white] svg{M86.3,186.2H70.9V79.1h15.4v48.4V186.2z}
        svg{M108.9,79.1h41.6c39.6,0,57,28.3,57,53.6c0,27.5-21.5,53.6-56.8,53.6h-41.8V79.1z M124.3,172.4h24.5c34.9,0,42.9-26.5,42.9-39.7c0-21.5-13.7-39.7-43.7-39.7h-23.7V172.4z}
        svg{M88.7,56.8c0,5.5-4.5,10.1-10.1,10.1c-5.6,0-10.1-4.6-10.1-10.1c0-5.6,4.5-10.1,10.1-10.1C84.2,46.7,88.7,51.3,88.7,56.8z};
    }
}

\newcommand\orcidicon[1]{\href{https://orcid.org/#1}{\mbox{\scalerel*{
                \begin{tikzpicture}[yscale=-1,transform shape]
                \pic{orcidlogo};
                \end{tikzpicture}
            }{|}}}}

\newcommand{\ORCIDDanny}{$^{\textsuperscript{\orcidicon{0000-0002-4538-7814}}}$}    
\newcommand{\ORCIDFrederik}{$^{\textsuperscript{\orcidicon{0000-0001-5482-9787}}}$} 
\newcommand{\ORCIDMichael}{$^{\textsuperscript{\orcidicon{0000-0001-6326-4749}}}$}  
\newcommand{\ORCIDTobias}{$^{\textsuperscript{\orcidicon{0000-0002-0682-4284}}}$}   

\begin{document}

\bstctlcite{disable_urls}

\title{Improving Automated Visual Fault Detection by Combining a Biologically Plausible Model of Visual Attention with Deep Learning}

\author{
	\IEEEauthorblockN{
		Frederik Beuth    \ORCIDFrederik,
		Tobias Schlosser  \ORCIDTobias,
		Michael Friedrich \ORCIDMichael, and
		Danny Kowerko     \ORCIDDanny
	}
    \IEEEauthorblockA{
		Junior Professorship of Media Computing, \\
		Chemnitz University of Technology, \\
		09107 Chemnitz, Germany, \\[0.4em]
		\begin{tabular}{ccc}
		 \texttt{\small frederik.beuth@cs.tu-chemnitz.de} & & \texttt{\small tobias.schlosser@cs.tu-chemnitz.de} \\
		 \texttt{\small michael.friedrich@cs.tu-chemnitz.de} & & \texttt{\small danny.kowerko@cs.tu-chemnitz.de }
		\end{tabular}
	}
}

\maketitle

\begin{abstract}
    It is a long-term goal to transfer biological processing principles as well as the power of human recognition into machine vision and engineering systems. One of such principles is visual attention, a smart human concept which focuses processing on a part of a scene. 
    In this contribution, we utilize attention to improve the automatic detection of defect patterns for wafers within the domain of semiconductor manufacturing. 
    Previous works in the domain have often utilized classical machine learning approaches such as KNNs, SVMs, or MLPs, while a few have already used modern approaches like deep neural networks (DNNs). However, one problem in the domain is that the faults are often very small and have to be detected within a larger size of the chip or even the wafer. Therefore, small structures in the size of pixels have to be detected in a vast amount of image data. 
    One interesting principle of the human brain for solving this problem is visual attention.
    Hence, we employ here a biologically plausible model of visual attention for automatic visual inspection. 
    On this basis, we propose a hybrid system of visual attention and a deep neural network.
    As demonstrated, our system achieves among other decisive advantages an improvement in accuracy from 81\,\% to 92\,\%, and an increase in accuracy for detecting faults from 67\,\% to 88\,\%. Therefore, the error rates are reduced from 19\,\% to 8\,\%, and notably from 33\,\% to 12\,\% for detecting a fault in a chip. Hence, these results show that attention can greatly improve the performance of visual inspection systems.
    Furthermore, we conduct a broad evaluation, which identifies specific advantages of the biological attention model in this application, and benchmarks standard deep learning approaches as an alternative with and without attention.
    
    This work is an extended arXiv version of the original conference article published in \enquote{IECON 2020}. It has been extended regarding visual attention, covering (i) the improvement and equations of visual attention model, (ii) a deeper evaluation of the model, (iii) a discussion about possibilities to combine the attention model with the DNN, and (iv) a detailed overview about the data.    

\end{abstract}


\input{content/1_Intro}
\input{content/2_Methods}

\input{content/3_Results}

\input{content/4_Discussion}
\input{content/5_Conclusion}

\bibliographystyle{IEEEtran}
\bibliography{library}

\input{appendix}

\end{document}

%% file: content/1_Intro.tex
\section{Introduction}
\label{sec:intro}

\thispagestyle{psp1}

\begin{figure*}[tb]
	\centering

	\input{TikZ/figWafer_v4}

	\caption{Overview of a wafer (left) with chips (middle) as well as faulty and fine streets (right) as a result of the wafer dicing process.  
	(The shown examples were generated synthetically from the original images to protect the intellectual property, while retaining a close resemblance to the original imagery).}
	\label{fig:wafer}
	\vspace*{-0.5em}
\end{figure*}
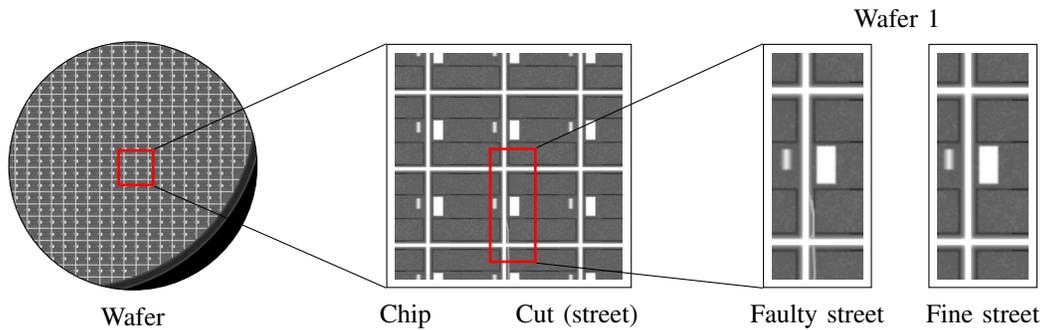

One long-term goal is to incorporate biological processing principles into machine vision systems. Visual attention, a smart human processing principle that focuses processing resources on an aspect of a scene relevant for the current task, is one of these biological processing principles \cite{Hamker2005b,Beuth2019}. We apply this principle here to the domain of wafer dicing 
to investigate its benefits and under which circumstances it improves automated visual inspection systems. The long-term goal of our research is therefore to better understand the power of human processing as well as to incorporate its benefits into machine vision systems and improve them accordingly. 

A major aim in the domain of the semiconductor industry is to detect and recognize production errors and faults early on. As manual detection is a very labor-intensive and thus costly procedure, computer vision systems are often deployed as an automatic detection system \cite{Huang2015a,Kumar2008}. This does not only results in reduced manufacturing costs and work load, but also helps increasing the yield of the production process itself. Hence, systems for automated visual inspection are widely deployed in the industry.

In this contribution, we address the topic of wafer dicing. Wafer dicing is the separation of silicon wafers into single components, e.g. chips, often using a dicing saw \cite{hooper2015,rahim2017}. 
Dicing based on laser technology is a novel alternative method to separate
brittle semiconductor materials via thermally induced mechanical forces (laser cut wafer dicing \cite{hooper2015,rahim2017}). Thereby, a dicing street is the area where dicing is potentially allowed. The quality criterion of dicing is that the laser cut (kerf) must not leave the street  (Fig. \ref{fig:wafer}). A curve leaving a street entails faulty chips and decreases the wafer yield, i.e. the ratio of faultless to the total sum of chips.

The field of automated visual inspection for wafer dicing uses classically image processing approaches commonly differentiated due to their functionality in projection-, filter-
based, and hybrid approaches \cite{Huang2015a, Schlosser2019}. Projection-based approaches
include for example principal component analysis, whereas filter-based approaches encompass spectral estimation methods, yet, they often need manual adaption. Therefore, learning-based and hybrid approaches make utilization of support vector machines (SVMs) or multilayer perceptrons (MLPs) \cite{Fei-LongChen2000, Xie2014}, while, in recent years, also a few more powerful deep learning (DL, \cite{LeCun2015}) approaches have been deployed \cite{Cheon2019, Lee2017, Lee2018, Nakazawa2018, OLeary2020}, which we will address here. For instance, \cite{Nakazawa2018} tested synthetic wafer data using a convolutional neuronal network (CNN) in their application. Others recognize the production process over time for an on-the-fly control \cite{Lee2017}, or use recurrent neuronal networks \cite{Lee2018}. 

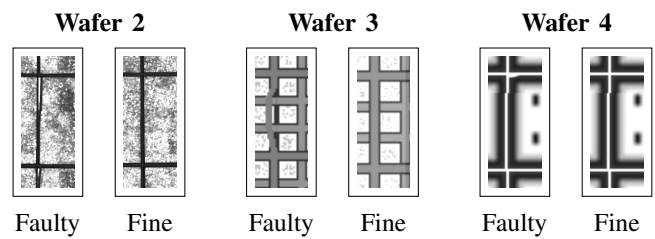
\begin{figure}
	\centering

    \hspace{-0.8cm}\input{TikZ/figWaferStreets}
	\caption{Our data material is heterogeneous, and it consists of several different wafers, for which exemplary streets are depicted.}
	\label{fig:wafer2}	
	\vspace*{-0.75em}
\end{figure}

However, one problem is that faults are typically very small in size and hard to detect in the large wafer disk (Fig. \ref{fig:wafer}). 
Imaging systems capture the complete (often stitched) image of a wafer, which results in resolutions of up to 150 megapixels, where single chips often range in a size of 200 -- 2000 pixels as in our case. Faults are even smaller structures up to the size of only a few pixels (Fig. \ref{fig:wafer}, \ref{fig:wafer2}). Therefore, the challenge is to detect these small structures in a vast amount of data. Classical deep neural networks have trouble to deal with this problem: Either the input image is down-scaled as normally for the network, but then the small structures would be lost. Or, the input could be chosen large enough, but then the network would run very slowly and had too many free parameters, leading to overfitting problems, as the network's size would increase unfeasible.

One interesting biological processing principle for this problem is visual attention, a smart human mechanism to select from the huge amount of input data the relevant one for the task at hand \cite{Carrasco2011}, or to focus neuronal processing resources on an aspect of a scene \cite{Hamker2005b, Beuth2019}. 
The principle is deployed here to ``zoom in'' into the wafer disk or into a chip (Fig. \ref{fig:wafer}).
Visual attention has, to the author's knowledge, not been used in the domain yet. Therefore, we like to propose the first model of visual attention,
in combination with deep learning, for the domain of fault recognition in the semiconductor industry.

There exist several approaches of combining visual attention with deep neural networks (DNNs) in other domains already. The background behind this idea is often that attention focuses the processing on a part or an aspect or a scene, which is then passed for classification to a machine learning classifier, like a DNN. In such a way, the attention model zooms in on the content, which then the DNN classifies in higher resolution and with less irrelevant data. However, the way to combine this remains pretty unclear and under debate. 
We found several dominant approaches, which are listed in the following.
1) One approach is the saliency models \cite{Itti1998}. The idea understands attention as a mechanism to define a spatial region (region of interest, ROI \cite{Itti1998}), which is then later passed to a classifier. Newer work uses this idea for taking fast snapshots from the image via attention and feed them to a CNN (Glimpse approach \cite{Ba2015}).
2) As saliency models struggle with objects that are not very obvious in the image, they find the wrong regions for the DNN. This has led to a combination of this approach with machine learning like reinforcement learning \cite{Sutton1998}, e.g. \cite{Cao2017}.
3) Similarly, a saliency model can be controlled by words to find the visual regions corresponding to text (attention for visual-textual alignment, e.g. Wang et al. \cite{Wang2017}).
4) A different approach is from Jürgen Schmidthuber \cite{Stollenga2014}, which introduces top-down feature-based attention in his model and thus uses attention towards specific features, not regions.
5) Attention towards some basic features can also be used to select an irregularly shaped region, defined via visual features instead of a spatial region, and then feed this content to a DNN (e.g. \cite{Zhao2018}). 
6) In the recent few years, works also utilize DNNs, or building blocks of them, as attention networks. So, one neuronal network serves as attention network, while the other is modulated by it and processes normally the image (e.g. \cite{Cai2019, Fu2019_CVPR, Zhao2018}). Quite a few systems use this approach, but they are designed very diverse (5 examples: \cite{Cai2019,  Chen2019_CVPR, Fu2019_CVPR, Guo2019_CVPR, Zhao2018}).

Therefore, many different approaches exist that allow the combination of visual attention with machine learning and deep learning based models, yet, selecting the approach with the best performance for a specific task remains difficult.

However, when we look at the attentional processing in the brain (e.g. Miller \& Buschman \cite{Miller2013} or Tsotsos et al. \cite{Tsotsos2005}) and the underlying connectivity (e.g. Ungerleider et al. \cite{Ungerleider2008} or Felleman \& Van Essen \cite{Felleman1991}), we found that none of the above approaches resembles closely the attentional processing in the human brain. Also none of these approaches are underpinned with much neuroscientific data. For instance, there exist a lot of single-cell recordings in the attention literature \cite{Reynolds2009, Beuth2015ax} which are not replicated at all, and also many behavioral influences by attention \cite{Carrasco2011}. The saliency models are inspired by the latter \cite{Itti1998}, but the approaches are nowadays also not so close more linked to psychology. Moreover, there exist also many other different ways in which visual attention alters human behavior, which are not reflected (e.g. texture segmentation \cite{Thielscher2008a,Qiu2007}).

Therefore, we propose here to use a more biologically plausible model of visual attention, to avoid all of these problems, and the quarrels about which model is the right one. 
Biologically-plausible models of visual attention originate more from the discipline of computational neuroscience, which develops models of the human brain to replicate neuronal recording data and to simulate human behavior. In recent years, a few works have also shown that such models of visual attention are capable of real-world applications (see dissertation \cite{Beuth2019} for an overview), e.g. \cite{Hamker2005b,Tsotsos2005,Antonelli2014,Beuth2019,Thielscher2008b,Chikkerur2010}. However, none of these models, at least to the authors' literature search, have been combined with deep learning yet. 

We will employ the model of Beuth, 2019 \cite{Beuth2019} as it shows very promising real-world capability (see also \cite{Beuth2015b}) while maintaining a deep biological plausibility. The model has a strong neurophysiological foundation as it can replicate a large range of neuronal recordings of attention \cite{Beuth2015ax}. Moreover, it has a satisfying operation as several applications show it can at least deal with virtual reality \cite{Jamalian2016} and real-world \cite{Beuth2015b}. And finally, it can replicate human behavior as it is based on previous models (visual search, \cite{Hamker2005a,Hamker2005b}) and is able to also fit new experiments in other behavioral paradigms (OSM, Chap. 5 in \cite{Beuth2019}). Additionally, the work \cite{Beuth2019} shows advantages over saliency models: a) Top-down object-descriptors, b) biological foundation, c) the option to easily realize complex task sets in a natural way, as later shown in Sec. \ref{sec:attModel}.
The current work is deeply rooted in neuroscience and focuses on the concept of visual attention, while extending an earlier work-in-progress publication of us using opposingly a specialized pipeline \cite{Schlosser2019}. The current biologically-grounded contribution also evaluates, in comparison to this shorter work, the concept of visual attention with deep learning more broader and in-depth. This is more thoroughly possible as we utilize now a model based on the human brain.
Therefore, we like to propose the combination of a biologically-plausible model of visual attention with deep learning as our second contribution.

%% file: TikZ/figWafer_v4.tex
\begin{tikzpicture}[c/.style={draw, circle}, r/.style={draw, rectangle}]
	\node[c, minimum size=3.3cm, path picture={\node at (path picture bounding box.center)
        {\includegraphics[width=3.3cm]{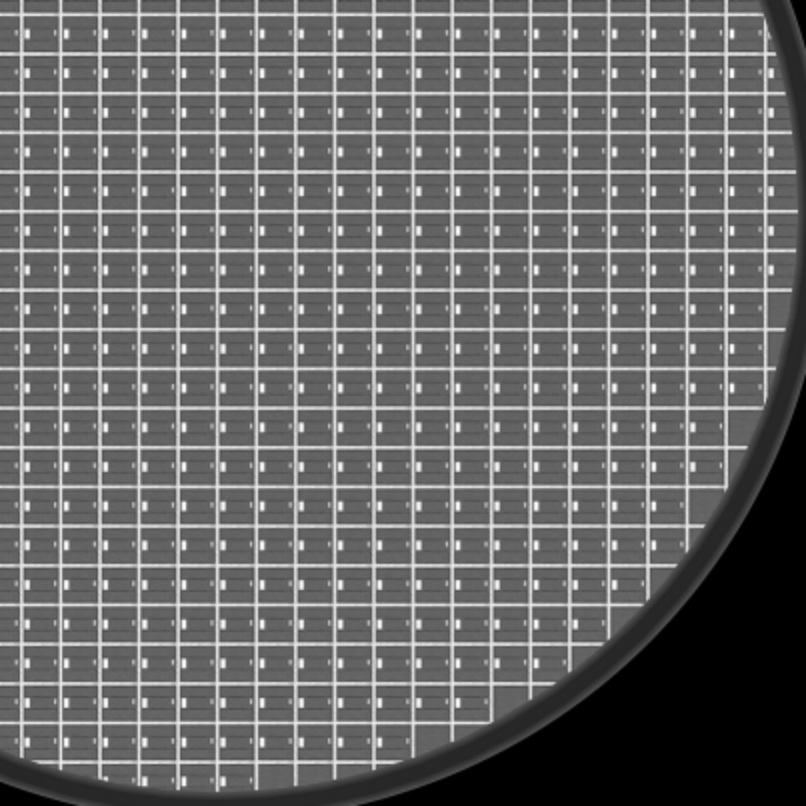}};}] (c) at (0, 0) {};
	\node[r, draw=red, line width=1pt, minimum size   = 0.46cm] (r2) at (0.04, -0.025) {};
	\node[r, minimum height = 3cm]    (r3) at ( 5,   0)    {\includegraphics[height=3cm]{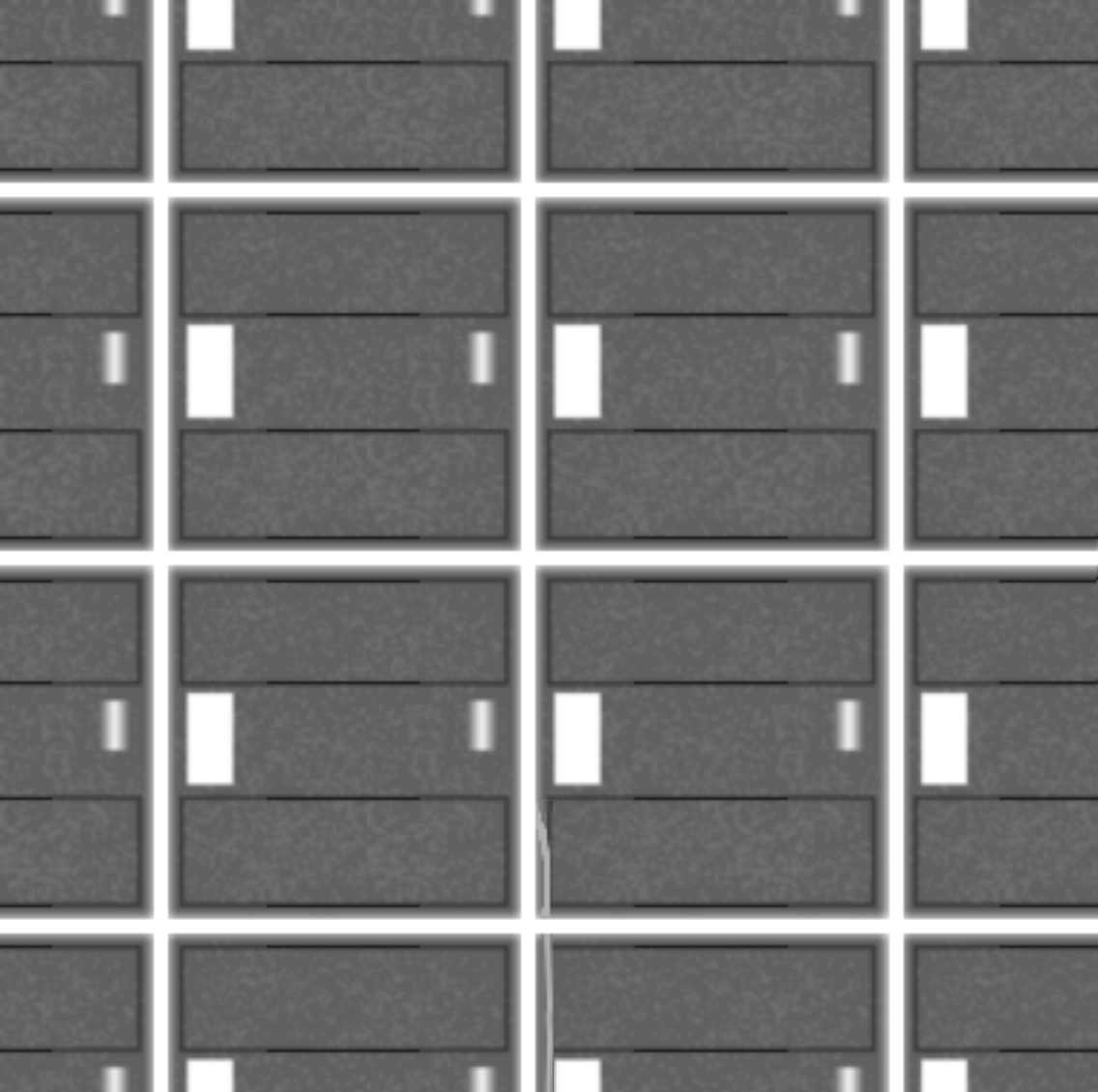}};
	\node[r, draw=red, line width=1pt, minimum width  = 0.6cm, minimum height = 1.5cm] (r4) at (5.05, -0.525) {};
	\node[r, minimum height = 3cm]    (r5) at ( 9.1, 0)    {\includegraphics[height=3cm]{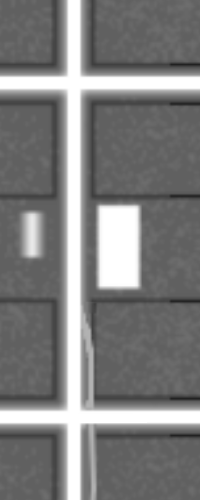}};
	\node[r, minimum height = 3cm]    (r6) at (11.3, 0)    {\includegraphics[height=3cm]{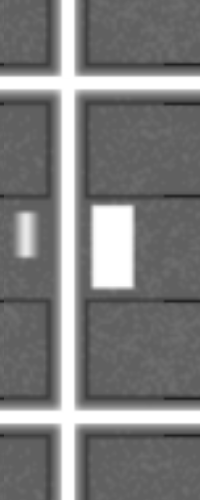}};

	\node[below=0.1cm of c]  (cl)  {Wafer};
	\node[below=0.1cm of r3] (r3l) {Chip\hspace{1cm} Cut (street)};
	\node[below=0.1cm of r5] (r5l) {Faulty street};
	\node[below=0.1cm of r6] (r6l) {Fine street};
    \node[above=0.1cm of r5]  (r_5_6_l)   {\hspace{2cm} Wafer 1};

	\draw (r2.north east) -- (r3.north west);
	\draw (r2.south east) -- (r3.south west);

	\draw (r4.north east) -- (r5.north west);
	\draw (r4.south east) -- (r5.south west);
\end{tikzpicture}

%% file: TikZ/figWaferStreets.tex
\scalebox{0.97}{\begin{tikzpicture}[c/.style={draw, circle}, r/.style={draw, rectangle}]

	\node[r, minimum height = 1.8cm] (r7)  at (0.0, 0) {\includegraphics[height=1.8cm]{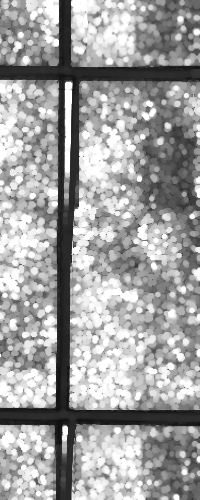}};
	\node[r, minimum height = 1.8cm] (r8)  at (1.4, 0) {\includegraphics[height=1.8cm]{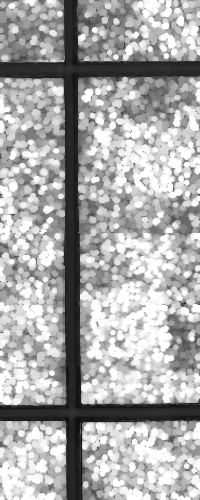}};
	\node[r, minimum height = 1.8cm] (r9)  at (3.2, 0) {\includegraphics[height=1.8cm]{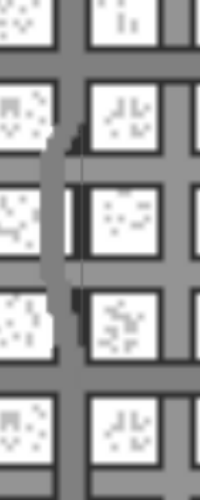}};
	\node[r, minimum height = 1.8cm] (r10) at (4.6, 0) {\includegraphics[height=1.8cm]{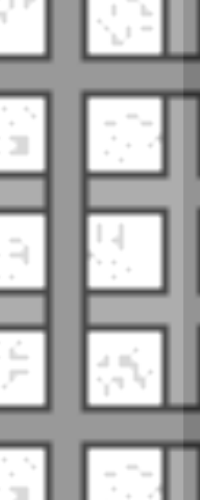}};
	\node[r, minimum height = 1.8cm] (r11) at (6.4, 0) {\includegraphics[height=1.8cm]{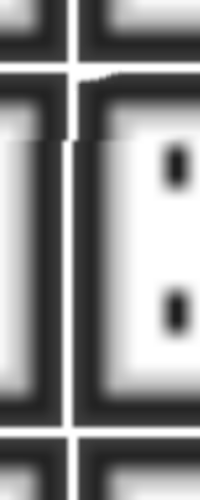}};
	\node[r, minimum height = 1.8cm] (r12) at (7.8, 0) {\includegraphics[height=1.8cm]{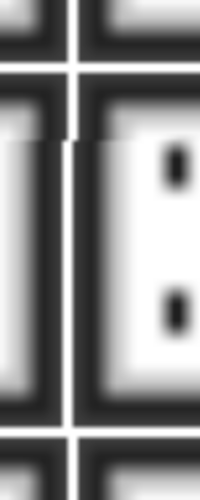}};

	\node[above=0.1cm of r7]  (r_7_8_l)   {\hspace{1.3cm} \textbf{Wafer 2}};
	\node[above=0.1cm of r9]  (r_9_10_l)  {\hspace{1.3cm} \textbf{Wafer 3}};
	\node[above=0.1cm of r11] (r_11_12_l) {\hspace{1.3cm} \textbf{Wafer 4}};

	\node[below=0.1cm of r7]  (r7l)  {Faulty};
	\node[below=0.1cm of r8]  (r8l)  {Fine};
	\node[below=0.1cm of r9]  (r9l)  {Faulty};
	\node[below=0.1cm of r10] (r10l) {Fine};
	\node[below=0.1cm of r11] (r11l) {Faulty};
	\node[below=0.1cm of r12] (r12l) {Fine};

\end{tikzpicture}}

%% file: content/2_Methods.tex
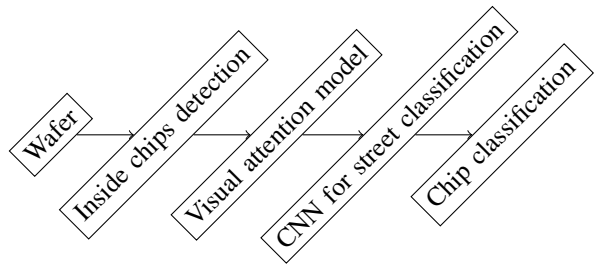
\begin{figure}[tb]
    \centering

    \input{TikZ/figSystem}

    \caption{Overview of the whole system.}
    \label{fig:system}
    \vspace*{-0.5em}
\end{figure}

\section{Proposed system}

We design a multi-stage system (Fig. \ref{fig:system}), which processes a whole wafer to detect faulty and faultless chips. Wafer images are recorded via different microscopes, either in the form of unstitched subimages or, alternatively,  preprocessed via the microscope software by stitching all subimages into one image. In the system, the chips are at first classified into chips inside the wafer area and chips on the wafer border. The latter are also scanned by the microscope, but are typically incomplete or broken (Fig. \ref{fig:wafer}, wafer border). The manufactures are not much interested in faults in the outer chips, thus we only process the inner chips further.

Afterwards, the visual attention model is employed to find the region of interest (ROI) for the inside chips. Regions of interest are in our application the chip borders, streets, and their surroundings. The ROIs are then passed to the CNN for detecting faults in the regions. 
Finally, we calculate if a chip is faulty or not, depending on the classification of the four borders of each chip.
To show the benefits of attention, we will compare our full system to approaches without attention and to approaches of the automated visual fault detection.

\subsection{Classification of inside vs. wafer-border chips}

This first stage distinguishes chips inside the wafer from chips on the wafer border. The border chips are typically broken so they have to be labeled for the user as ``wafer border chip'' and are excluded from evaluation. However, the recognition is pretty easy as the pattern is very obvious. The chip is crossed by a large black shape constituting the wafer border (Fig. \ref{fig:wafer}). We found, that a basic CNN is already able to achieve an accuracy of over $99\,\%$, utilizing two convolution layers, one max pooling, one dense and one fully connected layer.
This corresponds to a VGG network \cite{Simonyan2014} with only one block. 

\begin{figure}[t]
	\centering

	\includegraphics[width=0.97\linewidth]{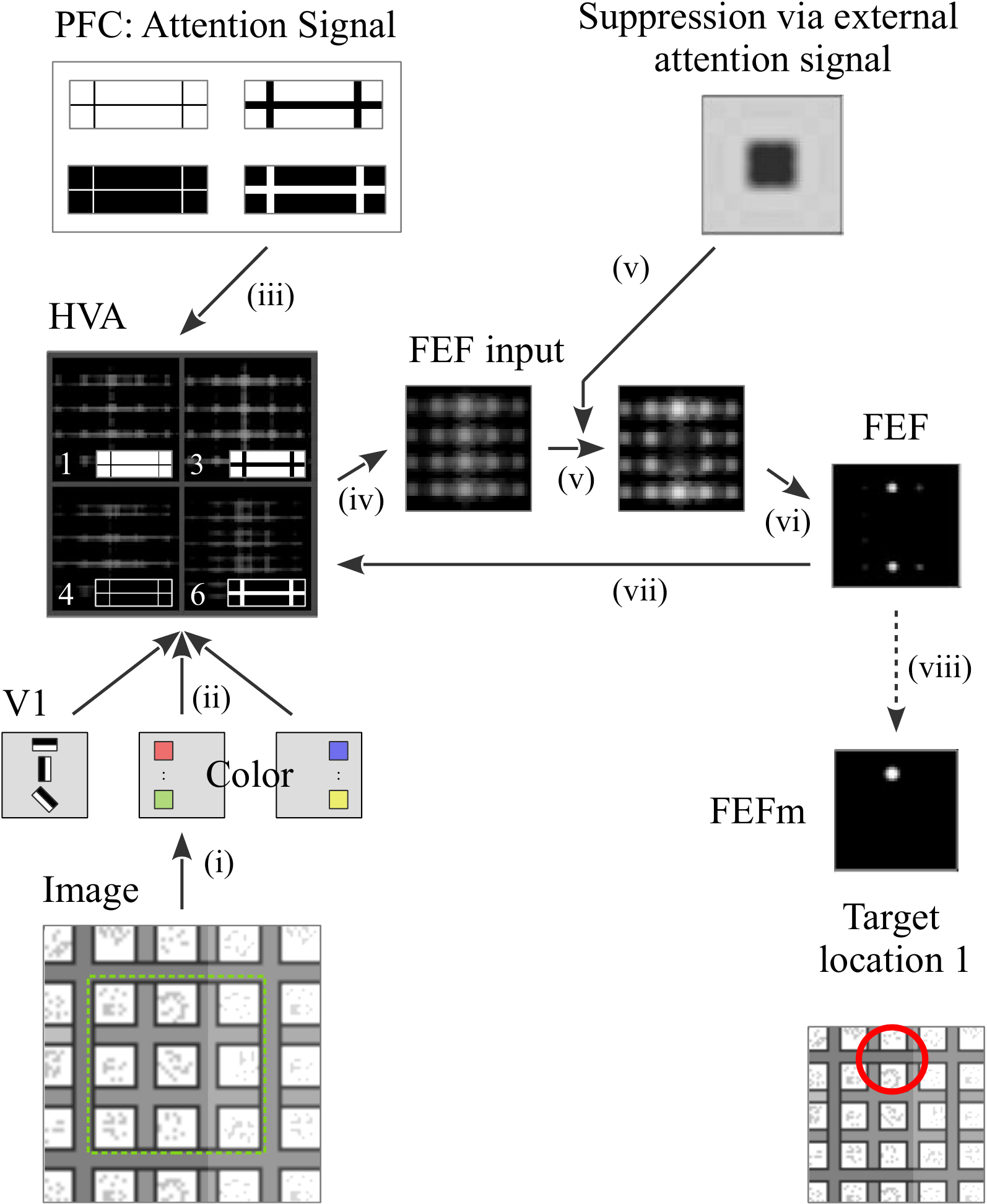}

	\caption{The visual attention model in the task to search for a street. The input image shows a chip of wafer 3, whereby again the image was abstracted. In the higher visual area (HVA), each box shows the activity of a street neuron in image coordinates, whereby what pattern the neuron detects is visible in the inlet. The red circle marks the center position of the found target. See the main text for the area abbreviations. The chip's borders are marked with a dotted green line.}
	\label{fig:attModelRun}
	\vspace*{-0.75em}
	
\end{figure}

\subsection{Model of visual attention}
\label{sec:attModel}

Our attention model (Fig. \ref{fig:attModelRun}) stems from a line of computational-neuroscience models of visual attention from the group of Fred Hamker \cite{Beuth2019,Jamalian2016,Beuth2015ax,Beuth2015b,Zirnsak2011,Hamker2005b}. The model is adapted and further developed from \cite[Chap. 4]{Beuth2019}. 

In the attention model, at first, the input image is processed through an earlier visual brain area (V1), which filters the image for edges and colors (step (i) in Fig. \ref{fig:attModelRun}). Edges are recognized by Gabor filters, and colors as a color contrast between red and green, and between blue and yellow \cite{Beuth2019}.  Colors are not used much in this wafer application, but are included for the generality of the model. This stage models brain areas like the lateral geniculate nucleus (LGN) and the primary visual cortex (V1).

The neuronal responses are then routed to a higher-level visual area (HVA), encoding objects (ii). This area simulates high-level visual areas in the brain such as the inferior temporal cortex, where these cells have been found \cite{Logothetis1995}. The responses of the model's HVA neurons are computed by convolving the neurons of V1 with a pre-learned weight matrix. The weight matrix can be learned by any offline learning method, we use a simple one which was already employed in previous research, the one-shot learning \cite{Beuth2019,Jamalian2016}. The procedure learns an object directly from a single image, leading to its name. We apply it here to learn cells reacting to the chip borders. 
We learned 12 different cells to account for the variance in the data material: We combine 2 street orientations (vertical and horizontal), 3 street widths, and 2 color schemes (black streets/white chips and vice versa). Four cells are exemplarily shown in Fig. \ref{fig:attModelRun}, no. 1, 3, 4 and 6.

The model has to search in this task for the borders of a chip (Fig. \ref{fig:attModelRun}). This task is called visual search in psychology and it is known that top-down visual attention is applied to the search target \cite{Wolfe1994}, realized in the model by signals from the prefrontal cortex (PFC), encoding this task instruction, to the higher visual area (iii). These signals amplify neurons encoding the edges in HVA, leading to higher activity at potential edges in the image. This realizes the human task instruction ``check the streets of the chip'', which we know from human workers is carried out by looking at the chip border regions (streets).

Afterwards, the HVA activities are spatially processed by a brain area called frontal eye field (FEF \cite{Beuth2019}, iv-vii). The FEF first takes the maximum over the HVA activity, the result shows activity on potential target object locations. Afterwards, a competition between places is applied and the activity is projected back to HVA, forming a reentrant loop (vii). This loop focuses neuronal activity to a single location over time (spatial attention). When a single location is selected in FEF movement (FEFm), it denotes an upcoming eye movement. In humans, the eye movement would select the target location, thus we simply read out it and define it as street center (viii).

Which of the four streets is selected is more or less random. After the first street is selected, the location is suppressed in the FEF by an inhibition-of-return standard approach (IOR, \cite{Hamker2005b}) via the external attention signal, so another street will be selected. This task is repeated 4 times. Finally, 4 street regions are cut out based on the found center coordinates of each street. 

The model was further extended from previous work for this application by: introducing the IOR concept from previous attention models \cite{Hamker2005b}, adding an external attention signal, and increasing the precision of the street center. The changes are listed in the Sec. \ref{sec:modelImprov}. In general, the model is described by a set of differential equations \cite{Beuth2019}.

\subsection{Advantages of the attention model}

Besides the general advantage of visual attention, we have analyzed our work regarding the advantages of such a biological attention model in the current application.

First, the model searches a street pattern based on high-level object descriptors and not merely pixel-based. The object is encoded by a neuron in the higher visual area (Fig. \ref{fig:attModelRun}). Multiple different streets can be encoded by multiple object templates, thus also a certain variance can be encoded. The model then searches for all templates in parallel. The high-level object descriptors make the model more robust compared to classical computer vision approaches like edge following, e.g. against noisy areas or stitching errors, while keeping the model simpler than deep learning solutions.

Due to the nature of the employed learning, i.e. the one-shot learning, the method learns a template directly from a single image. Hence, it is very fast in its nature and the total runtime of the learning is only a few seconds on consumer-grade hardware. 
The other advantage of the learning procedure is that the image can be defined in a rather conceptual manner, more of a sketch than an actual image (Fig. \ref{fig:attModelRun}). We found that even defining the image via simple image editor software is enough. Hence, it can be easily and swiftly produced.

As a last advantage, the attention model can deal with the inner structures of the chips out of the box by an external spatial attention signal. This signal defines which chip areas are suppressed regarding processing. We use it to suppress the inner structure of a chip, as we know human inspectors would also not look at the middle of a chip image. To illustrate this, we use a chip with especially a lot of inner structures in Fig. \ref{fig:attModelRun}. The suppression map can again simply be defined as an image. The model naturally realizes in this way task instructions. During the processing, the initially very noisy activity is filtered, illustrated by showing the input stages towards the FEF (iv,v,vi): the first stage is very noisy as it reacts to all inner structures of the chip (iv). Yet, the external spatial attention signal is then applied to this, which suppresses the ``middle'' by decreasing neuronal activity (v). Afterwards, the competition takes place, increases the signal contrast, and reduces the activity to a few locations (vi). During this processing, the model filters out inner chip structures, which might be very similar to the searched streets and thus would divert the recognition process.  

\input{content/2_Model_description.tex}

\begin{table}[t]
    \renewcommand{\arraystretch}{1.1}
    \centering

    \scalebox{0.85}{\begin{tabular}{|M{1.25cm}|M{1.5cm}|M{1.5cm}|M{1.9cm}|M{0.8cm}|M{0.8cm}|}
        \hline
		Unit & Layer & Type & Output shape & Kernel size & Stride \\
        \hline
        \noalign{\vskip 2pt}

        \hline
        \multirow{4}{*}{conv1} & conv1\_1 & conv & $56\times188\times32$ & $5 \times 5$ & $1$ \\
        \cline{2-6}
        & conv1\_2 & conv & $54\times186\times48$ & $3 \times 3$ & $1$ \\
        \cline{2-6}
        & pool1 & max pool & $18\times62\times48$ & $3 \times 3$ & $3$ \\
        \cline{2-6}
        & dropout1 & dropout & $18\times62\times48$ & / & / \\
        \hline
        \multirow{4}{*}{conv2} & conv2\_1 & conv & $16\times60\times64$ & $3 \times 3$ & $1$ \\
        \cline{2-6}
        & conv2\_2 & conv & $14\times58\times96$ & $3 \times 3$ & $1$ \\
        \cline{2-6}
        & pool2 & max pool & $7\times29\times96$ & $2 \times 2$ & $2$ \\
        \cline{2-6}
        & dropout2 & dropout & $7\times29\times96$ & / & / \\
        \hline
        \multirow{4}{*}{conv3} & conv3\_1 & conv & $5\times27\times144$ & $3 \times 3$ & $1$ \\
        \cline{2-6}
        & conv3\_2 & conv & $3\times25\times192$ & $3 \times 3$ & $1$ \\
        \cline{2-6}
        & pool3 & max pool & $3\times8\times192$ & $1 \times 3$ & $1 \times 3$ \\
        \cline{2-6}
        & dropout3 & dropout & $3\times8\times192$ & / & / \\
        \hline
        \multirow{3}{*}{fully conn} & dense1 & fully conn & $192$ & / & / \\
        \cline{2-6}
        & dropout4 & dropout & $192$ & / & / \\
        \cline{2-6}
        & dense2 & fully conn & $2$ & / & / \\
        \hline
    \end{tabular}}

    \caption{The CNN for street classification.}
    \label{tab:cnnStructureStreets}
    \vspace*{-0.5em}
\end{table}

\subsection{CNN for street classification}
\label{sec:methodCNNstreet}

The streets were then classified by a convolutional network, as outlined in Tab. \ref{tab:cnnStructureStreets}, in 3 classes. 'Good streets' (Class 0)
represents chip regions without a fault, and 'bad streets' (2) with a fault. Additionally, we defined a class 'anomalies'
(1), describing intact streets, but with an unknown visual event on them. The plan is to report back these chips in
the later production system to a human worker for a further manual inspection. For the analyses in this paper, this class
is not considered, and its samples are put into the 'good streets'.

The attention model returns the center where it has found a street. We define a region of interest (ROI, \cite{Itti1998}) of $120\,\%$ the size of the chip and $6 \times$ the width of the street around this center. The goal of our application is to detect dicing faults, which occur in the space between the streets and the chips, and possibly continue inside the chip. To cover these areas at best, the ROI was centered in such a way that the main street is placed at $\sfrac{1}{3}$ of the image height such that $\sfrac{2}{3}$ of the image shows the associated chip. Hence, the input street image contains the street itself as well as parts of adjacent chips and street crossings (Fig. \ref{fig:wafer}, right side). 

Regarding the design of the CNN, we indicate for the CNN the relevant chip areas inside the provided image and thus what part of the image should contain no cuts, by employing a trick from deep reinforcement learning \cite{Mnih2013}. The street regions are rotated before being passed on to the CNN, whereas the chip area is always located in the same image position (chosen as top). Thus, the position of the chip region is stable over all images in the data set. The trick is now to reduce spatial pooling to allow the CNN to learn specific weights for specific regions in the provided street image.
As the distinction lays predominantly in the $y$ direction (chip is on top, street on bottom),  we lower the pooling in the direction of $y$. 
Following this principle, the effect of the pooling in $y$ is reduced within the deeper pooling layers as shown in Table \ref{tab:cnnStructureStreets}.

\subsection{Chip classification}

Finally, the results of the street classification are translated back to determine faulty and good chips. If a chip has at least one faulty side, it is defined as faulty, otherwise, it is defined as good. The chip classification is the final outcome of the system.

%% file: TikZ/figSystem.tex
\begin{tikzpicture}[r/.style={draw, rectangle, rotate=45}]
	\node[r] (l1) at (0,   0) {Wafer};
	\node[r] (l2) at (1.5, 0) {Inside chips detection};
	\node[r] (l3) at (3,   0) {Visual attention model};
	\node[r] (l4) at (4.5, 0) {CNN for street classification};
	\node[r] (l5) at (6,   0) {Chip classification};

	\draw[->] (l1) -- (l2);
	\draw[->] (l2) -- (l3);
	\draw[->] (l3) -- (l4);
	\draw[->] (l4) -- (l5);
\end{tikzpicture}

%% file: content/2_Model_description.tex
\input{content/mathEnv.tex}

\subsection{Improvements of the visual attention model}
\label{sec:modelImprov}

The attention model has been modified in several ways for this work.
Additionally, as the attention model stems from a line of older attention models, this section provides the changes as compared to the previous works \cite{Beuth2019}.

\subsubsection{Spatial external attention signal}

The model was extended with an external spatial attention signal. In the data set, some of the chips have line-like structures within the chip. We know from human workers that they were told to ignore the structures inside a chip, which translates roughly to the task instruction `do not look in the middle of a chip'. Such task instructions can be naturally realized in the model by directing `negative' attention to the middle of a chip. 

Hence, this signal realizes spatial components of task instructions. We implemented this external attention signal as an incoming signal to the frontal eye field (FEF), as this structure is highly involved in spatial attention \cite{Pouget2009,Hamker2005a}.

The equations in the FEF were changed in the FEF visual cells from the ones in \cite{Beuth2019} to the following ones by adding the term of $\left[\cdot \left(1+2\,a\Area{FEF}\right)\right]$:
\begin{eqnarray}
    \tau\Area{FEFv} \, \frac{\partial r\Area{FEFv}_{x}}{\partial t} &=&  - r\Area{FEFv}_{x}  + E_{x} \\
    \label{eq:eFEF} \mbox{with: } E_{x} &=& C \left( Q \left(F_x\right) \right) \\
    \label{eq:FEF_F} F_x&=&  \normZeroOne{E\Area{HVA2}_{x} \cdot \left(1+2\,a\Area{FEF}_{x}\right)}
\end{eqnarray}
whereby $a\Area{FEF}$ denotes the spatial attention signal, resulting in a neuronal activity matrix of the same size as the FEF. The entries are $a\Area{FEF}_x \in [-0.25, 0.25]$ and thus can be negative to allow a `negative' attention signal. The variable $E\Area{HVA2}_{x}$ denotes the incoming signal from the HVA. 
In the doctoral thesis \cite{Beuth2019}, the FEF received also an incoming signal from V1 ($E\Area{V1}_{x}$), which is disabled here as it is not used and thus set to $0$ (equations number Eq. 4.53 - 4.55 in \cite{Beuth2019}). The helper functions $C(x)$ and $Q(x)$ denote a non-linearity ($C$) to increase the difference between low and high signals (Eq. 4.59 in \cite{Beuth2019}, not listed here), and a signal enhancement operation ($Q$, Eq. 4.60 in \cite{Beuth2019}).

\subsubsection{Inhibition of return (IOR)}

To detect multiple streets in the chip image, an inhibition-of-return approach (IOR \cite{Hamker2005b,Itti1998}) was also added to the model. IOR describes the concept that the target region is inhibited after an eye movement, so the next eye movements do not visit this region again. This allows the search and localization of several streets in a row within a chip image. The IOR is implemented in our model by an IOR map, which stores locations of previous eye movements as activity blobs and suppresses then the FEF cells (standard approach, \cite{Hamker2005b,Itti1998}).

The idea was realized as follows: We define an inhibition map ($r\Area{IOR}$), initialized initially with a small positive number uniformly ($0.25$), representing no inhibition.
The image is then processed through the normal duration of the model. When an eye movement (saccade) is executed, inhibition at this position $x_s$ is set as a blob of negative values (Eq. \ref{eq:iorUp1}, \ref{eq:iorUp2}):
\vspace{-0.25em}
\begin{eqnarray}
     \label{eq:iorUp1} G &=& \mathfrak{g}(x_s,1,\frac{\#FEF_{x,y}}{6}) \\
    \label{eq:iorUp2} r\Area{IOR}_x &=& r\Area{IOR}_x - v\Area{IOR} \cdot G,
\end{eqnarray}

whereby:

\begin{itemize}
  \item The parameter $v\Area{IOR}=0.75$ denotes a scaling of the IOR influence.
  \item The function $\mathfrak{g}$ represents a two-dimensional Gaussian function centered at $(0,0)$, whereby $a$ denotes the amplitude, and $\sigma$ the standard deviation: \\[-0.5em] 
  $$\mathfrak{g}\left(x,a,\sigma \right) =  a \cdot \exp \left(-\left(\frac{(x_1-0)^2}{2\sigma^2_1} + \frac{(x_2-0)^2}{2 \sigma^2_2}\right) \right)$$
\end{itemize}
Afterwards, the image is processed again, and the map $r\Area{IOR}$ inhibits the FEFv cells on the previous locations. This inhibition leads to the selection of a different target, i.e. a different street. The IOR influence is also realized as a modulatory influence into the FEF, like a `negative' attention signal, and is bundled with the other attention signal, the external one, via a Fuzzy-Min operation \cite{Carpenter1992}. With the IOR component, the Eq. \ref{eq:FEF_F} changes to the following Eq. \ref{eq:FEF_F_IOR}:
\vspace{-0.25em}
\begin{equation}
    \label{eq:FEF_F_IOR} F_x=  \normZeroOne{E\Area{HVA2}_{x} \cdot (1+2 \cdot \min\left\{a\Area{FEF}_x, r\Area{IOR}_x\right\})}
\end{equation}

\subsubsection{Precision of the attention model's localization}

First of all, we increased the spatial precision of the attention model by three improvements: 

A) Adapting V1. V1 consists of two layers, V1 simple and pool, whereby the V1 pooling layer has a lower spatial resolution ($1:10$).
We improved the behavior of the model by aligning the two layers on top of each other, implying that V1 pool must have precisely $\sfrac{1}{10}$ of the size of V1 simple.

B) The behavior of the soft-max pooling in HVA layer 2/3 was changed in such a way that it considers high activities in the input (HVA layer 4) more strongly. HVA consists of two layers, HVA layer 4 and HVA layer 2/3, simulating the layers of a single cortical brain area \cite{Beuth2015ax}, whereby layer 2/3 is a pooling layer. For this change, the non-linearity in the soft-max operation was adapted from $p1=4, p2=1/4, v\Area{HVA4}=1 $ ~to~ $p1=8, p2=1/4, v\Area{HVA4}=16$ (Eq. \ref{eq:eL2}, corresp. to Eq. 4.50 in \cite{Beuth2019}).
\vspace{-0.25em}
\begin{eqnarray}
    \label{eq:eL2} E_{d,i,x} &=& \left( v\Area{HVA4} \cdot \sum \limits_{x' \in \text{RF}} w\Area{HVA4-HVA2}_{x'} \, (r\Area{HVA4}_{d,i,x'})^{p_1} \right)^{p_2} \\[0.25em] 
    \label{eq:wFFL2} w\Area{HVA4-HVA2}_{x'} &=& \mathfrak{g}(x',1,[1,1])
\end{eqnarray}

C) The size of HVA layer 2/3 and the FEF was increased to have a finer resolution for selecting the target. Normally, HVA layer 2/3 has a lower resolution than HVA layer 4 to decrease the spatial information over the visual layer hierarchy. However, as also the FEF has the same resolution as HVA layer 2/3 (assumption in the original model), we increased the resolution in this area to have also a higher resolution in the FEF, hence resulting in more accurate eye movements.

\subsubsection{Equations adapted to the task}
Furthermore, several additional parameters were adapted. They were changed as the attention model was applied to a new application input, which causes different strong and broad activities in the neuronal layers.

\begin{itemize}
    \item The signal HVA layer 4$\rightarrow$HVA layer 2/3 was not only increased due to the changed non-linearity (see previous paragraph, 3-B), but also to account for a general lower HVA layer 4 activity: $v\Area{HVA4}=1 \rightarrow 16$. 
    \item The signal FEFvm$\rightarrow$HVA layer 4 was decreased to account for a broader FEF activity:  $v\Area{FEFvm-HVA4}=4 \rightarrow 3$ and $v\Area{SP-1}=0.85 \rightarrow 0.3$ (Eq. 4.37, 4.44 in \cite{Beuth2019}).
\end{itemize}

%% file: content/mathEnv.tex
\newcommand{\Area}[1]{^{\mbox{\tiny #1}}}
\newcommand{\excRF}[2]{\text{excRF}\left(#1,#2\right)}
\newcommand{\RF}[2]{\text{RF}^{#1}_{x'}\left(#2\right)} 
\newcommand{\RFin}[1]{{{x'} \in \text{RF}\left(x,#1\right)}}
\newcommand{\mean}[2]{\text{M}_{#1}\left(#2\right)}
\newcommand{\pos}[1]{\left[#1\right]^+}
\newcommand{\posT}[1]{\left[#1\right]^+}
\newcommand{\normZeroOne}[1]{\left[#1\right]_0^1}
\newcommand{\mdeg}{^{\circ}}

%% file: content/3_Results.tex
\section{Test results and evaluation}

In this evaluation, we analyze the performance of our whole system, and quantify the improvements which result from employing visual attention. Yet, beforehand, we will validate if and how well the attention model operates in this task.

\subsection{Data overview and test configuration}

The wafer data set originates from a real-world, laser-based dicing process of multiple types of semiconductor wafers. 
During the process, each wafer was first mounted on a tape and a frame. Subsequently, the dicing tape was expanded in order to broaden the cut, making it possible to visualize the dicing streets using a wide-field light microscope. The used microscopes have a scanning stage which scans the wafer line-wise over n lines and m columns, allowing imaging of up to 150~mm~/ 6~inch wafers. The recorded subimages are stored separately or are stitched together, and then the images are saved offline. 

The data set consists of 10 different wafers as more wafer material was not available due to copyright restrictions. 
A more detailed breakdown of our data set is shown in Tab.~\ref{tab:waferNumbers}. 
The wafers belong to 6 different types and thus are pretty heterogeneous (Fig. \ref{fig:wafer}, \ref{fig:wafer2}). 
It is visible from the data that the samples per wafer are seriously imbalanced as the wafers exhibit different sizes and thus different numbers of chips in reality. This all must be considered by a real-world machine learning program.

\begin{table*}[t]
	\renewcommand{\arraystretch}{1.1}
	\centering

	\begin{tabular}{|c|c|c|c|c|c|c|c|c|c|c||c|}
		\hline
		Wafer & 1 & 2 & 3 & 4 & 5 & 6 & 7 & 8 & 9 & 10 & Total \\
		Wafer type  & 1 & 2 & 3 & 3 & 3 & 3 & 1 & 4 & 5 & 6 &  \\
		\hline
		\noalign{\vskip 2pt}

		\hline
		No. chips        & 744 & 5\thinspace 007 & 566 & 566 & 566 & 566 & 664 & 144 & 728 & 2\thinspace 050 & 11\thinspace 601 \\
		\hline
		No. inside chips & 550 & 3\thinspace 344 & 432 & 434 & 430 & 432 & 536 & 112 & 586 & 1\thinspace 336 &  8\thinspace 192 \\
		\hline
		\noalign{\vskip 2pt}

		\hline
		Class 0 -- good    & 2\thinspace 017 & 12\thinspace 150 & 1\thinspace 682 & 1\thinspace 539 & 1\thinspace 556 & 1\thinspace 638 & 1\thinspace 882 & 327 & 1\thinspace 789 & 5\thinspace 003 & 29\thinspace 583 \\
		\hline
		Class 1 -- anomaly & 90 & 239 & 27 & 143 & 144 & 64 & 151 & 78 & 421 &  12 & 1\,369 \\
		\hline
		Class 2 -- bad     & 90 & 554 & 15 &  51 &  18 & 23 & 110 & 39 & 109 & 313 & 1\,322 \\
		\hline
	\end{tabular}

    \caption{Number of samples in the data set, split into the numbers for each wafer, and for chips inside the wafer area. Additionally, the number of streets are given per wafer and per class.} 
	\label{tab:waferNumbers}
\end{table*}

\begin{table*}[t]
	\renewcommand{\arraystretch}{1.1}
	\centering

	\begin{tabular}{|c|c|c|c|c|c|c|c|c|c|c|c|c|}
		\hline
		Wafer       & 1 & 2 & 3 & 4 & 5 & 6 & 7 & 8 & 9 & 10 & Total \\
		\hline
		\noalign{\vskip 2pt}

		\hline
		Total streets  & 2\,200 & 13\,376 & 1\,728 & 1\,736 & 1\,720 & 1\,728 & 2\,144 & 448   & 2\,344 & 5\,344 & 32\,768 \\
		\hline
		Found streets  & 2\,197 & 12\,943 & 1\,724 & 1\,733 & 1\,718 & 1\,725 & 2\,143 & 444   & 2\,319 & 5\,328 & 32\,274 \\
		\hline
		Accuracy in \% & 99.86  & 96.76   & 99.77  & 99.83  & 99.88  & 99.83  & 99.95  & 99.11 & 98.93  & 99.70  & 98.49   \\ 
		\hline
	\end{tabular}

	\caption{Accuracy of the visual attention model to find the streets, given per each wafer.}
	\label{tab:segAccComplete} 
\end{table*}

Training of the CNNs was performed with the deep learning framework Keras and TensorFlow in Python. We split the data into a 50\,\% training, 25\,\% validation, and 25\,\% test set. All data were resized to $192 \times 60$ in advanced for the streets (and to $96 \times 96$ for the chips). We choose to have a rather small input size to keep the number of weights, i.e. free parameters, low, as our data set is rather small (Table \ref{tab:waferNumbers}).

As we have a very low amount of data for a deep-learning problem, we chose also to design our network rather small (Tab. \ref{tab:cnnStructureStreets}). Additionally, the data were augmented on-the-fly using the methods: randomized rotation in a range of up to $4^{\circ}$ as well as scaling up to $\pm 4\,\%$, translation in $x$~/ $y$ up to $10$~/ $1\,\%$, and flip in $x$.
As mentioned above, we would like to reduce the level of augmentation in the direction of $y$ on the chip border regions, thus we decrease the amount of translation in $y$ and omit the flip in $y$.
Finally, the data were contrast-normalized before being processed by the framework. 

The classes in the problem are very imbalanced (ratio is $92.2\,\%$ : $3.7\,\%$ : $4.1\,\%$, Tab. \ref{tab:waferNumbers}), hence they need to be balanced for the CNN. For this, we first try to weight the loss function higher when a class is less presented in the data set, but we found that this approach gives suboptimal accuracies for such a high imbalance. Hence, we tried a different approach and found that simply duplicating the samples for the underrepresented classes while applying data augmentation yields good results. 
The on-the-fly data augmentation ensures, despite several images are identical in the data set, that the images appear differently for the CNN after the augmentation. Therefore, such an approach produces a good amount of image variations on the fly from one source image \cite{Zhao2018}. 

\subsection{Evaluation of the visual attention model}

At first, we will verify if the attention model shows plausible human behavior. 
As we do not posses human eye tracking data for this task, and as there is, according to our literature search, no eye tracking data available for workers performing fault recognition in the semiconductor industry, we can only analyze if the neuronal activities and eye movements of the model are reasonable under a task like this and are in line with the general literature about visual search \cite{Wolfe1994,Hamker2005a,Wolfe2010}.
For this, we first illustrated the model's behavior on a representative example, as already shown during the model's description. This serves as a qualitative analysis. Afterwards, we quantify the results, i.e. the correctness of eye movements, which connotes here how many streets the attention model finds and how precise are the selected coordinates. 

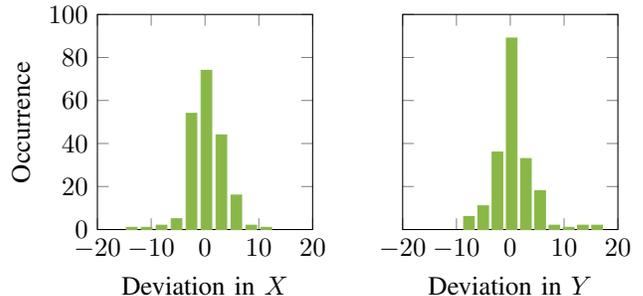
\begin{figure}[t]
	\centering

	\vspace*{-1.25em}
    \input{TikZ/figHistRegionPrecision}
	\caption{Precisions for the region of interest selection of the attention model. The precision is measured in pixels via the distance between the true center to the determined one.}
	\label{fig:histRegionPrecision}
	\vspace*{-0.5em}
\end{figure}

\subsubsection{Correctness of the found streets}
At first, we quantify the number of streets found. The model of visual attention finds $98.49\,\%$ of the existing $32\thinspace 768$ streets in our data set correctly, whereby the false positives were $29$ of these. The localization precision per each wafer is given in Tab. \ref{tab:segAccComplete}. We defined a street as not found if the extracted center coordinates are not on the chip sides ($0.3 < x, y < 0.7$) or if the ROI borders are outside the image borders. 
The data set consists of $8\,192$ processable chips, and thus has a total of $32\,768$ street segments, based on the assumption that each chip has four street segments. 
The model's correctness of $98.49\,\%$ is within the upper range of the shown accuracies in different tasks by previous literature. The first versions published in 2005 have achieved an accuracy of $50\,\%$\,\cite{Hamker2005a,Hamker2005b}, whereas later ones have reached in a rather easy task with 3 objects $96.2\,\%$\,\cite{Antonelli2014}. More recent versions have accomplished $92\,\%$ in a larger and more complex real-world application with 100 objects \cite{Beuth2019,Beuth2015b}, and $85.4\,\%$ in a different virtual reality application \cite{Jamalian2016}. Therefore, we conclude from this comparison that the model shows a very good performance to select the streets correctly via eye movements.

\subsubsection{Precision of the region of interest extraction}

To ensure a high-quality region of interest (ROI) extraction, we measured the centers' accuracies of the resulting street segments, whereas the street's center should be situated in the center of the image. These results were obtained by manually annotating the street centers of 200 randomly selected samples and measuring the pixel-based distance to the real image center.
The attention model achieves in average a deviation in $X$ and $Y$ of $0.34$ and $0.57$ pixels respectively (Fig.~\ref{fig:histRegionPrecision}), with corresponding standard deviations of $3.16$ and $3.65$ pixels. Humans typically make an eye movement within the vicinity of an object, and another eye movement(s) to fine-center the gaze. The second operation is not part of the model, but despite this, the model finds the ROIs quite accurately with mean deviations within the range of a few pixels. Hence, the model is probably more precise than as expected from human psychological findings.

In summary, the model shows reasonable behavior as expected during a visual search task like this, and achieves a high accuracy to localize the streets correctly.

\subsection{Performance of the CNN for street classification}

We compare here the performance of the full system with a system where visual attention was disabled.

\subsubsection{Performance of the CNN with visual attention}

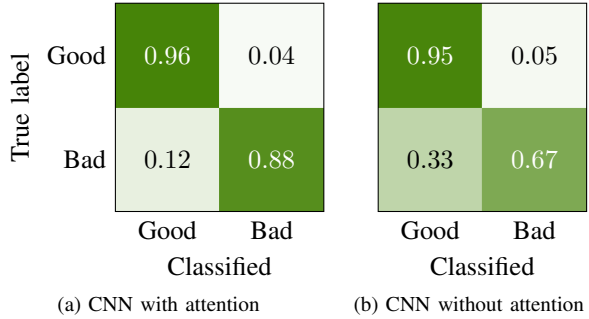
\begin{figure}[t]
	\centering

	\input{TikZ/figConfCnnGoodBad}

	\caption{Normalized confusion matrices depicting for a given chip (or street) on the $y$-axis how the system classified this chip ($x$-axis). The classification rates are given as percentages. \textbf{a)} Confusion matrix of the CNN with the visual attention model beforehand. \textbf{b)} Confusion matrix of the CNN without visual attention.}
	\label{fig:confMatrices}
	\vspace*{+0.5em}
\end{figure}

The full system achieves an accuracy to classify streets correctly of $91.91\,\%$ (Tab. \ref{tab:acc1}, mean accuracy over all classes). This and all other following values were measured over 5 runs. Additionally, we obtain a confusion matrix (Fig. \ref{fig:confMatrices}a). 
We are especially interested in the precision to detect faults on the wafer, and this accuracy is observable from the confusion matrix, showing faults are detected correctly with $87.80\,\%$ (also listen in Tab. \ref{tab:acc1}).

\begin{table}[t]
    \newcommand*{\tstsstac}{\thinspace\textsuperscript{\textasteriskcentered}}
    \newcommand*{\tsstac}{\textsuperscript{\textasteriskcentered}}

	\centering
    
    \begin{tabular}{|M{3.3cm}|M{2cm}|M{2cm}|}
		\hline
		Approach & Accuracy [\%] & Fault detection \\ &&accuracy [\%] \\
		\hline

		\multicolumn{3}{|c|}{Baseline} \\
		\hline
		KNN                           & $70.52 \pm 1.21$		& $55.00$ \\
		SVM                           & $78.33 \pm 1.34$ 		& $64.40$ \\ 
		MLP	                          & $65.26 \pm 5.85$ 		& $63.40$ \\ 
		CNN \cite{Nakazawa2018}       & $75.24 \pm 1.88$ 	   	& $57.20$ \\ 
		CNN \cite{Cheon2019}          & $78.89 \pm 3.03$ 		& $62.20$ \\ 
		CNN \cite{OLeary2020}         & $78.47 \pm 0.69$ 		& $60.20$ \\ 
        Our CNN                       & $80.83 \pm 2.38$   	    & $67.40$ \\
		\hline
    
        \multicolumn{3}{|c|}{Attention-based} \\
		\hline
		Attention + KNN	                      & $76.91 \pm 1.49$    &   $57.80 $ \\
		Attention + SVM	                      & $85.43 \pm 0.91$  	&   $74.60 $ \\
		Attention + MLP	                      & $79.85 \pm 3.52$ 	&   $73.25 $ \\
		Attention + CNN \cite{Nakazawa2018}   & $88.66 \pm 0.88$ 	&   $80.40 $ \\
		Attention + CNN \cite{Cheon2019}      & $87.63 \pm 1.20$ 	&   $76.80 $ \\
		Attention + CNN \cite{OLeary2020}     & $84.86 \pm 1.82$ 	&   $70.60 $ \\
		\textbf{Attention + Our CNN}          & $\mathbf{91.91 \pm 0.57}$ & $\mathbf{87.80}$ \\
		\hline

	\end{tabular}

	\caption{
    Street and chip classification results. Mean accuracies and standard deviations for (i) baseline solutions, and (ii) attention approaches based on our model.
	The CNNs \cite{Nakazawa2018}, \cite{Cheon2019}, \cite{OLeary2020} denote the state-of-the-art in the wafer domain.}
	\label{tab:acc1}
	\vspace*{-1em}	
\end{table}

\begin{table*}[t]
    \newcommand*{\tstsstac}{\thinspace\textsuperscript{\textasteriskcentered}}
    \newcommand*{\tsstac}{\textsuperscript{\textasteriskcentered}}

	\centering

	\begin{tabular}{|M{3.66cm}|M{1.9cm}|M{1.9cm}|M{1.9cm}|M{1.9cm}|M{1.9cm}|M{1.9cm}|}
		\hline
		\multirow{3}{*}{Approach} & \multicolumn{2}{c|}{Without attention} & \multicolumn{2}{c|}{With attention} & \multicolumn{2}{c|}{Improvement of} \\
		\cline{2-7}
		& Accuracy [\%] & Fault detection accuracy [\%] & Accuracy [\%] & Fault detection accuracy [\%] & Accuracy & Fault detection accuracy \\
		\hline
		\noalign{\vskip 2pt}

		\hline
		DenseNet121~\cite{LeCun2015}         & $77.47 \pm 1.57$ & $ 56.80 \pm 2.95$ & $84.86 \pm 1.38$ & $ 71.00 \pm 2.12$ & $7.39$ & $14.20$ \\
    	DenseNet121\tsstac~\cite{LeCun2015}  & $80.47 \pm 0.97$ & $ 62.80 \pm 2.49$ & $84.33 \pm 1.40$ & $ 69.20 \pm 2.95$ & $3.86$ & $ 6.40$ \\
    	InceptionV3~\cite{LeCun2015}         & $77.18 \pm 1.35$ & $ 57.20 \pm 3.03$ & $83.08 \pm 1.57$ & $ 66.80 \pm 3.49$ & $5.90$ & $ 9.60$ \\
    	InceptionV3\tsstac~\cite{LeCun2015}  & $76.27 \pm 1.76$ & $ 54.40 \pm 3.97$ & $82.42 \pm 2.68$ & $ 65.60 \pm 5.32$ & $6.15$ & $11.20$ \\
    	MobileNetV2~\cite{LeCun2015}         & $71.08 \pm 1.01$ & $ 44.40 \pm 2.51$ & $83.83 \pm 1.04$ & $ 68.40 \pm 2.07$ &$12.76$ & $24.00$ \\
        MobileNetV2\tsstac~\cite{LeCun2015}  & $76.99 \pm 1.36$ & $ 56.20 \pm 3.03$ & $84.19 \pm 1.10$ & $ 69.00 \pm 2.24$ & $7.21$ & $12.80$ \\
    	ResNet50~\cite{LeCun2015}            & $75.95 \pm 1.80$ & $ 55.60 \pm 4.77$ & $84.04 \pm 1.37$ & $ 69.00 \pm 3.00$ & $8.09$ & $13.40$ \\
    	ResNet50\tsstac~\cite{LeCun2015}     & $79.01 \pm 2.41$ & $ 59.80 \pm 5.22$ & $84.77 \pm 0.86$ & $ 70.40 \pm 1.82$ & $5.76$ & $10.60$ \\         
    	VGG10~\cite{LeCun2015}               & $83.19 \pm 1.64$ & $ 71.00 \pm 4.69$ & $86.75 \pm 2.08$ & $ 74.60 \pm 4.45$ & $3.57$ & $ 3.60$ \\
        VGG13~\cite{LeCun2015}               & $82.68 \pm 2.24$ & $ 69.80 \pm 5.07$ & $86.30 \pm 1.94$ & $ 73.80 \pm 3.96$ & $3.63$ & $ 4.00$ \\
        Xception~\cite{LeCun2015}            & $77.27 \pm 0.52$ & $ 57.40 \pm 0.89$ & $83.99 \pm 0.90$ & $ 68.60 \pm 2.07$ & $6.71$ & $11.20$ \\
        Xception\tsstac~\cite{LeCun2015}     & $80.52 \pm 0.74$ & $ 62.80 \pm 1.79$ & $85.18 \pm 0.88$ & $ 70.60 \pm 1.82$ & $4.66$ & $ 7.80$ \\
	    \textbf{Our CNN} & $80.83 \pm 2.38$ & $ 67.40 \pm 7.57$ & $\mathbf{91.91 \pm 0.57}$ & $\mathbf{87.80 \pm 1.92}$ & $11.08$ & $20.4$ \\
    	\hline
	\end{tabular}

	\caption{Performance evaluation of classical deep neural networks from computer vision. Tested once as standard solutions without attention, and once when combined with our attention approach. The last columns show the improvement through attention. \textsuperscript{\textasteriskcentered}Transfer learning, pretrained on ImageNet. } 
	\label{tab:acc2}
\end{table*}

\subsubsection{Performance without visual attention}

To highlight the benefits of visual attention, we benchmark a system where the attention model was removed from the pipeline (Tab. \ref{tab:acc1}), and add for comparision also other methods from the domain of wafer dicing as baseline. 
The CNN in this system uses directly whole chip images as input. This CNN reaches a mean accuracy of $80.83\,\%$ (Tab.~\ref{tab:acc1}). This accuracy does not look so bad, however, the accuracy to detect a fault correctly is only $67.40\,\%$. (Fig. \ref{fig:confMatrices}b). This value is much lower showing the effect of the visual attention model. 
In fact, the overall error rates increase without attention from $8\,\%$ to $19\,\%$, and for the faulty chips from $12\,\%$ to even $33\,\%$. These values show that especially the detection of faulty chips benefits from visual attention. 

Next, we evaluated our system in the context of related as well as DL-based systems in the domain of automated visual inspection for semiconductor manufacturing and wafer dicing (Tab.~\ref{tab:acc1}). While as of the current state-of-the-art automated visual inspection systems often ML-based approaches are being deployed, i.e KNN-, SVM-, and MLP-based classifiers, we include those approaches in our evaluation to provide a more comprehensive comparison. Besides them, the newest DNNs are tested \cite{Nakazawa2018, Cheon2019, OLeary2020}.
This does not include  DNNs that focus on e.g. process monitoring over time due the required time component \cite{Lee2017}, and contributions which do not provide sufficient details for a reimplementation \cite{Lee2017,Lee2018}.

Our evaluation shows that the proposed visual attention incorporating system constitutes a notable improvement over existing solutions, in which single CNN based systems \cite{Nakazawa2018, Cheon2019, OLeary2020} constitute the current state-of-the-art. 
While the CNNs of \cite{Nakazawa2018,Cheon2019} are the current state-of-the-art and reported in their publication accuracies of $98.2\,\%$ and $96.2 \,\%$ on their problems respectively, they reach here merely $75.24\,\%$ and $78.89 \,\%$. This implies for us our problem could be more challenging. A look at their data material confirms this, their data is more homogeneous and with much larger and thus easier to detect faults.  
In summary, we conclude that the proposed system is, at least from published approaches, currently the best system for this problem.

\subsubsection{Standard deep networks and benefits of attention}

In the next investigation, we benchmarked different standard DL-based approaches to (i) evaluate how they perform in comparison and thus could be used as alternative, and (ii) how much visual attention benefits them. While conventional DNNs are broadly applied to different real-world tasks, yet, they were also mainly developed for the recognition of real-world objects (e.g., ImageNet). Thus, the question arises how they perform in our use case with very different test data.
Table~\ref{tab:acc2} illustrates the resulting accuracies for different DNN-based models and architectures, showing our solution surpasses them too.

We found that all networks show a performance boost by visual attention, hence all networks benefit from visual attention. This illustrates that attention is a crucial principle and promotes its idea of zooming in. Surprisingly, the standard networks do not cope with attention very well, seen as the accuracies for the streets are lower than with our customized network. That implies the standard deep networks profit less from attention. 
Furthermore, transfer learning \cite{LeCun2015} benefits even less from attention.
We suppose that is because in the original data set (ImageNet), the objects are not shown in high resolution or large sizes like the faults in our street regions, hence the image material does not transfer well enough to our use case. 
Hence, our conclusion here is that visual attention is able to show its strengths much better if it is not mixed with transfer learning. This is reasonable since the data material is different.

Therefore, we conclude that our proposed system improves and outperforms, as evaluated in comparison, current state-of-the-art solutions in the wafer dicing domain.
Other standard deep neural network approaches show also lower accuracy percentages, and we evaluate generally the benefits of visual attention.
\enlargethispage{-\baselineskip}

\begin{figure}
	\centering

    \includegraphics{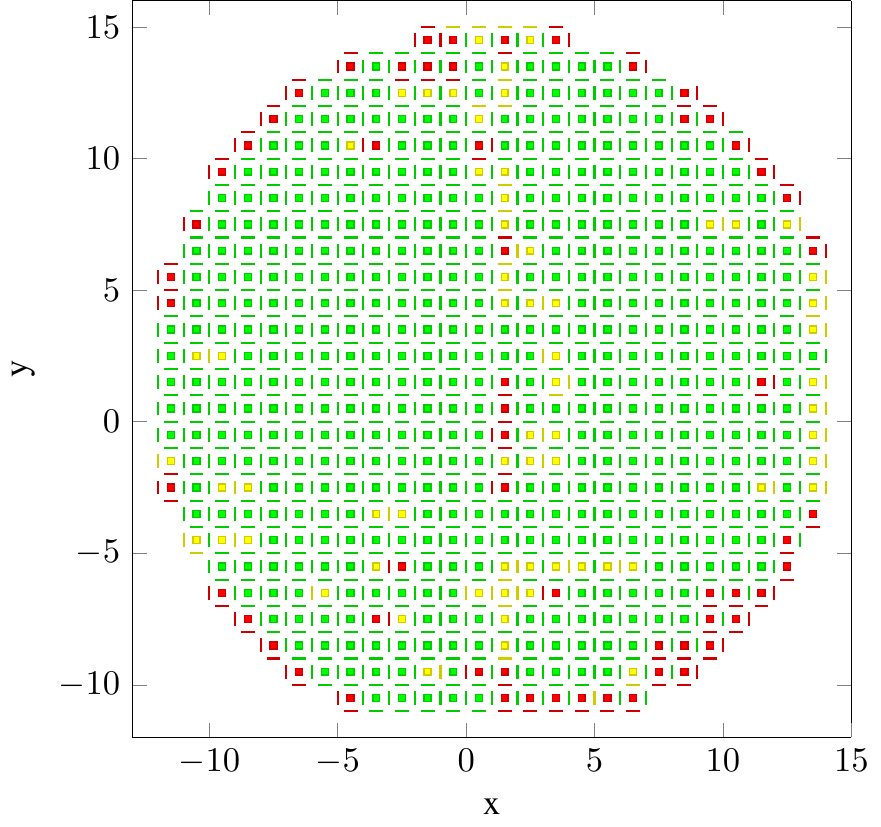}
    
	\caption{Street and chip classification ground truth visualized for good (\textcolor{green}{\textbullet}), anomaly (\textcolor{yellow}{\textbullet}), and faulty (\textcolor{red}{\textbullet}) streets and chips.}
	\label{fig:waferOverview}
\end{figure}

\subsection{Performance of the whole system}

\enlargethispage{-\baselineskip}
Finally, the class of a chip is calculated from the classification of its four street regions. The full system reaches the following accuracy to classify the chips correctly: $91\,\%$. 
This rate would be the important one for a final production system. 
The result of the classified streets and chip
error classes can then be illustrated, here as ground truth, as in Fig. \ref{fig:waferOverview}. It comprises the street and chip classification test results for good (\textcolor{green}{\textbullet}), anomaly (\textcolor{yellow}{\textbullet}) and faulty (\textcolor{red}{\textbullet}) streets and chips according to the used addressing scheme of the wafer. The shown error classes can then continue to differ in their respective defect pattern to be further assessed by an inspector.

%% file: TikZ/figHistRegionPrecision.tex
\subfloat{
\begin{tikzpicture}
	\begin{axis}[xmin=-20, xmax=20, ymin=0, ymax=100, width=0.245\textwidth, height=0.245\textwidth, xlabel={Deviation in $X$}, ylabel={Occurrence}, bar width=4pt]
		\addplot+[ybar, LimeGreen!90!black, fill=LimeGreen!90!black, mark=none] table[col sep=comma] {data/A_7_v2_a.csv};
	\end{axis}
\end{tikzpicture}}
\quad
\subfloat{
\begin{tikzpicture}
	\begin{axis}[xmin=-20, xmax=20, ymin=0, ymax=100, width=0.245\textwidth, height=0.245\textwidth, xlabel={Deviation in $Y$}, bar width=4pt, yticklabels={}]
		\addplot+[ybar, LimeGreen!90!black, fill=LimeGreen!90!black, mark=none] table[col sep=comma] {data/A_7_v2_b.csv};
	\end{axis}
\end{tikzpicture}}

%% file: TikZ/figConfCnnGoodBad.tex
    \subfloat[CNN with attention]{
        \begin{tikzpicture}
            \begin{axis}[
             xlabel={Classified}, ylabel={True label},
             xtick style={draw=none}, xtick={0, 1}, xticklabels={Good, Bad}, ytick style={draw=none}, ytick={0, 1}, yticklabels={Bad, Good},
             width=0.24\textwidth, height=0.24\textwidth,
             view={0}{90}, point meta min=0, point meta max=1, enlargelimits=false, axis on top,
             colormap={summap}{color=(white) color=(LimeGreen!50!black)},
             nodes near coords, nodes near coords align={center},
             nodes near coords black white/.style={
                 small value/.style={text=black},
                 large value/.style={text=white},
                 every node near coord/.style={/pgf/number format/fixed, check for zero/.code={
                     \pgfmathfloatifflags{\pgfplotspointmeta}{0}{\pgfkeys{/tikz/coordinate}}{
                	 \begingroup\pgfkeys{/pgf/fpu}\pgfmathparse{\pgfplotspointmeta<#1}
                	 \global\let\result=\pgfmathresult\endgroup\pgfmathfloatcreate{1}{1.0}{0}
                	 \let\ONE=\pgfmathresult
                	 \ifx\result\ONE
                	     \pgfkeysalso{/pgfplots/small value}
                	 \else
                	     \pgfkeysalso{/pgfplots/large value}
                	 \fi
                 }}, check for zero}}, nodes near coords black white=0.5]
                \addplot[matrix plot*, point meta=explicit] file[meta=index 2] {data/confMatrix_withAttention.dat};
            \end{axis}
        \end{tikzpicture}}
    \quad
    \subfloat[CNN without attention]{
        \begin{tikzpicture}
            \begin{axis}[
             xlabel={Classified},
             xtick style={draw=none}, xtick={0, 1}, xticklabels={Good, Bad}, ytick style={draw=none}, yticklabels={},
             width=0.24\textwidth, height=0.24\textwidth,
             view={0}{90}, point meta min=0, point meta max=1, enlargelimits=false, axis on top,
             colormap={summap}{color=(white) color=(LimeGreen!50!black)},
             nodes near coords, nodes near coords align={center},
             nodes near coords black white/.style={
                 small value/.style={text=black},
                 large value/.style={text=white},
                 every node near coord/.style={/pgf/number format/fixed, check for zero/.code={
                     \pgfmathfloatifflags{\pgfplotspointmeta}{0}{\pgfkeys{/tikz/coordinate}}{
                	 \begingroup\pgfkeys{/pgf/fpu}\pgfmathparse{\pgfplotspointmeta<#1}
                	 \global\let\result=\pgfmathresult\endgroup\pgfmathfloatcreate{1}{1.0}{0}
                	 \let\ONE=\pgfmathresult
                	 \ifx\result\ONE
                	     \pgfkeysalso{/pgfplots/small value}
                	 \else
                	     \pgfkeysalso{/pgfplots/large value}
                	 \fi
                 }}, check for zero}}, nodes near coords black white=0.5]
                \addplot[matrix plot*, point meta=explicit] file[meta=index 2] {data/confMatrix_withoutAttention.dat};
            \end{axis}
    \end{tikzpicture}}

%% file: content/4_Discussion.tex
\section{Discussion about how to couple a visual attention model with a CNN}

Several options exist how to couple a visual attention model with a CNN as delineated in the introduction of this work, e.g. \cite{Ba2015, Cai2019, Cao2017, Wang2017, Stollenga2014, Zhao2018}. 
The classical approach is that attention selects a spatial region (region of interest, ROI \cite{Itti1998}), 
which is subsequently passed to a classifier for later recognition. 
This idea came up at first with the saliency models in the late 1990s, which select a ROI for a later classifier \cite{Itti1998}. The classifier can of course today also be a CNN. Variations of this approach are the glimpse approach from Google \cite{Ba2015}, where a more modern neural network selects regions very swiftly (glimpses), which are then fed into a CNN.

Alternatives to the ROI approach would be other developments like proto-objects, where regions are formed like an object \cite{Walther2006}. They eliminate the drawback of the ROI approach that the region has always the shape of a rectangle, and hence may not fit a more naturally shaped object. 

Another alternative would be that visual attention modulates the neuronal response (amplify or suppress), which is closer to the processing in the brain \cite{Reynolds2009}. 
This approach starts to be utilized in the deep learning community to a small degree (e.g. \cite{Zhao2018}).
The idea is to feed all area into the CNN, but enhance neuronal responses in the attended region, and suppress all neurons in other places. The approach has the advantage that it can shape the region more clearly. Additionally, the amplification and suppression is closer to the neuronal processing of the brain (\cite{Reynolds2009, Beuth2015ax}, original data e.g. \cite{Reynolds1999}), because attention in the brain operates multiplicatively such as multiplicatively enhancing or suppressing neurons.

We consider also to utilize this novel modulation approach, but finally decided against it:

i) The modulation approach requires to feed entire images and not only the ROIs into the CNN. As the images are relatively large in our application (up to 1\,000~pixels across), this would result in a large CNN and hence in a reduced processing speed, as well as, in a lot of free parameters and overfitting (we have only a few data here). On the other hand, the street-showing ROIs in our case have a resolution of only $400 \times 60$ pixels, which is many times smaller than the entire chip image.

ii) 
The advantage of the newer approach is that the region can be shaped more clearly and flexibly. However, this big gain is in our wafer application of not much benefit, as the most structures in our image material are anyway rectangles. 

iii) 
Despite the approach's closer similarity to the brain's functionality, the neuroscientific review literature also reveals that the effects of attention are much more complex than a simple `enhance attended' or `suppress all others' \cite{Beuth2015ax,Reynolds2009}. As we aim for a strong biological plausibility, we would not like to ignore these effects lighthearted, and rather aim for a separate research publication investigating how a deep neuronal network can be integrated into an attention model or vice versa in a biologically plausible way.

iv) Finally, we have a complicated task-instruction here. The workers' instruction is to check the cuts for potential faults, which is of course realized by looking at the cuts (i.e. streets). Hence, we have applied attention to the streets here. 
However, not a street alone need to be checked for cracks, but rather also the close region next to it, ranging from the close surrounding of the street over the street-chip border towards the inside of the chip. 
Therefore, the workers' task requires to transfer from the attended structure (streets) to another spatial area. As it is not really known how such task instructions are realized in the brain, we do not know how to implement this in the attention model. The region of interest approach instead solves this problem out-of-the-box.

Therefore, we evaluate it is better to use the region of interest approach, which solves the problem nicely and has advantages in our case. 

%% file: content/5_Conclusion.tex
\section{Conclusion and outlook}

In this contribution, we propose a novel system for automated visual fault detection by combining a biologically-plausible model of visual attention with a deep neural network. The process of automated visual fault detection in the domain of semiconductor manufacturing and laser-based wafer dicing constitutes one particularly challenging application area, as defect patterns often range within a size of only a few pixels\,/\,{\textmu}m.
This problem is challenging for traditional convolutional neuronal networks, and it is getting more challenging due to the heterogeneity and imbalance of the image material. 
Visual attention is well suited for this problem, but not much used in the semiconductor industry yet, for which we created the first deep learning system with attention. Our benchmark shows that visual attention improves the mean accuracy from $81\%$ to $92\%$, and the accuracy to detect faults correctly from $67\%$ to $88\%$. Hence, the error rate especially for the faults drops from $33\%$ to $12\%$. These rates outperform notable other state-of-the-art systems in the domain, as well as the power of standard deep learning systems. 

This work utilizes a biologically-plausible model that is deeply rooted in neuroscience \cite{Beuth2019, Beuth2015ax, Hamker2005b, Zirnsak2011} for the combination of visual attention with deep learning approaches. As of the current state of the art, it is often unclear how both principles from visual attention and learning-based approaches should be combined in an application, furthermore they often lack biological plausibility and therefore tend to be unreliable or outperformed (Sec. \ref{sec:intro}). 

Future projects can be built on top of this system and can be enhanced with it, for example in other domains. The zooming-in approach is certainly not limited to the inspection of wafer faults, many other mechanical engineering problems may exist where small faults have to be recognized in a large amount of data as well as also surely in domains outside mechanical engineering. 
Moreover, future work can cover improvement for series production, or an application-specific optimization of the imagery system. The imagery system is currently standard and can be certainly enhanced and optimized for the current application.

\section*{Acknowledgement}

This work was partially funded by German Federal Ministry of Education and Research (BMBF) within project group localizeIT, project OphthalVis 2.0, and the ESF.

%% file: appendix.tex
\section{Appendix}

In the first part of the appendix, we will list the CNN without visual attention. It utilizes as input directly whole chip images, and classifies them into good and faulty chips.
In the second part, we provide a wafer overview and the street and chip groundtruth for each wafer. 

\subsection{CNN without visual attention}

When we compare our system to a system without attention, we remove the attention model. Hence, the deep neural network has to process whole chips, resulting in a new CNN (Tab. \ref{tab:cnnStructureChips}). We designed for a fair comparison both networks as similar as possible. However, as chips are squared and not rectangle-shaped anymore, and as the trick with reduced spatial pooling in one dimension is no longer necessary (Sec. \ref{sec:methodCNNstreet}), the last stage contains a normal pooling stage and the pooling size has to be changed to $3 \times 3$ (Tab. \ref{tab:cnnStructureChips}). 
The squared CNN's input size was chosen to hold the same amount of input pixels, at least roughly. The street CNN has an input shape of $60\times192 = 11520$ pixels. This result in a squared size of $\sqrt{11520} = 107$ pixels. We round it to the next multiple of 32, to make it feasible for the warp-size in the GPU, resulting in the finally chosen size of $96\times96$. Otherwise, the networks are the same. 
To verify that the size does not inflict any side effects, we run in addition a verification test with a bigger size of $192\times192$ and found no differences in the accuracy.

\subsection{Insight into our data material}

We received ten different wafers from a semiconductor manufacturer with different sizes, faults, materials, and imaging conditions. In the following, we have displayed all individual wafers. After analyzing the data, we created schematic overviews of each wafer to illustrate the distribution of faults by showing for each wafer which chips and streets are good and which have faults. Additionally, anomalies are marked that represents intact streets, but with an unknown visual event on them. The first wafer is shown in the main result section (Fig. \ref{fig:waferOverview}), and the remaining ones in this appendix (Fig. \ref{fig:waferOverviewAll1} - \ref{fig:waferOverviewAll5}) as the figures are relatively large in size and detail. These schematic diagrams illustrate the distribution of faults and give an impression, along with the wafers overviews, about the size and shape of each wafer for the reader. The wafer images were again scaled down and slightly modified to protect the intellectual property of the company.

\setlength{\@fptop}{0pt}

\begin{table}[t]
    \renewcommand{\arraystretch}{1.1}
    \centering

\scalebox{0.85}{\begin{tabular}{|M{1.25cm}|M{1.5cm}|M{1.5cm}|M{1.9cm}|M{0.8cm}|M{0.8cm}|}
        \hline
		Unit & Layer & Type & Output shape & Kernel size & Stride \\
        \hline
        \noalign{\vskip 2pt}

        \hline
        \multirow{4}{*}{conv1} & conv1\_1 & conv & $92\times92\times32$ & $5 \times 5$ & $1$ \\
        \cline{2-6}
        & conv1\_2 & conv & $90\times90\times48$ & $3 \times 3$ & $1$ \\
        \cline{2-6}
        & pool1 & max pool & $30\times30\times48$ & $3 \times 3$ & $3$ \\
        \cline{2-6}
        & dropout1 & dropout & $30\times30\times48$ & / & / \\
        \hline
        \multirow{4}{*}{conv2} & conv2\_1 & conv & $28\times28\times64$ & $3 \times 3$ & $1$ \\
        \cline{2-6}
        & conv2\_2 & conv & $26\times26\times96$ & $3 \times 3$ & $1$ \\
        \cline{2-6}
        & pool2 & max pool & $13\times13\times96$ & $2 \times 2$ & $2$ \\
        \cline{2-6}
        & dropout2 & dropout & $13\times13\times96$ & / & / \\
        \hline
        \multirow{4}{*}{conv3} & conv3\_1 & conv & $11\times11\times144$ & $3 \times 3$ & $1$ \\
        \cline{2-6}
        & conv3\_2 & conv & $9\times9\times192$ & $3 \times 3$ & $1$ \\
        \cline{2-6}
        & pool3 & max pool & $4\times4\times192$ & $2 \times 2$ & $2$ \\
        \cline{2-6}
        & dropout3 & dropout & $4\times4\times192$ & / & / \\
        \hline
        \multirow{3}{*}{fully conn} & dense1 & fully conn & $192$ & / & / \\
        \cline{2-6}
        & dropout4 & dropout & $192$ & / & / \\
        \cline{2-6}
        & dense2 & fully conn & $2$ & / & / \\
        \hline
    \end{tabular}}    

    \caption{Layer configuration of the convolutional neuronal network for chip classification.}
    \label{tab:cnnStructureChips}
\end{table}

\vfill
\clearpage
\input{TikZ/figWafer_Overviews}

%% file: TikZ/figWafer_Overviews.tex
\begin{figure*}[p]
	\centering

    \subfloat{\includegraphics[width=0.35\linewidth]{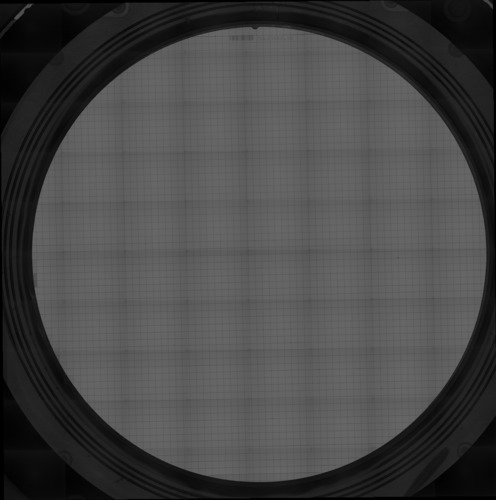}}
    \quad
	\subfloat{\includegraphics[width=0.45\linewidth]{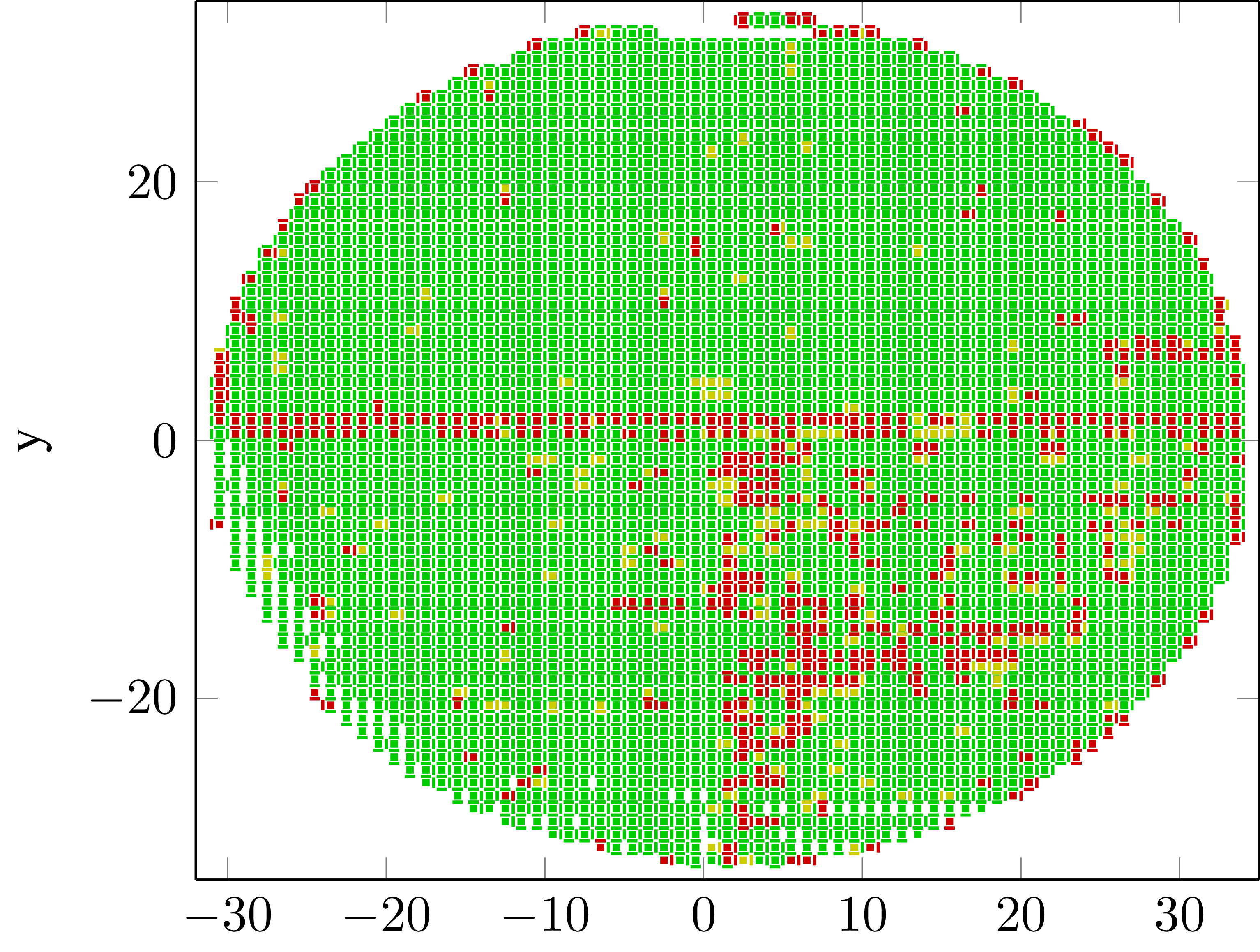}}

    \subfloat{\includegraphics[width=0.35\linewidth]{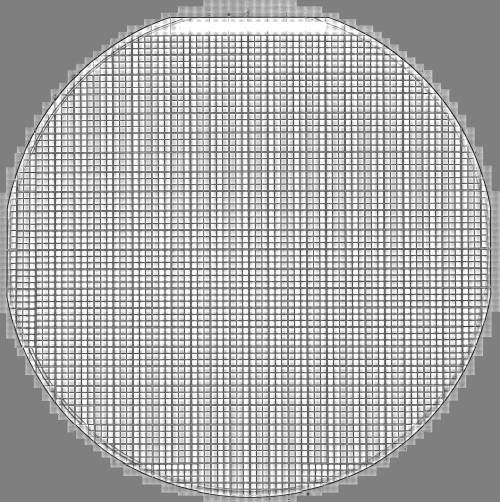}}
    \quad
	\subfloat{\includegraphics[width=0.45\linewidth]{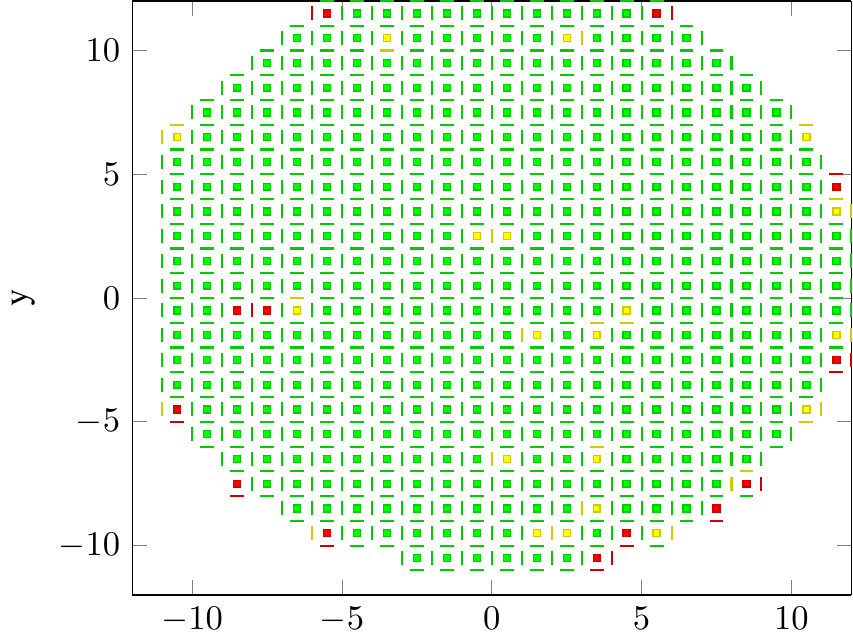}}

    \subfloat{\includegraphics[width=0.35\linewidth]{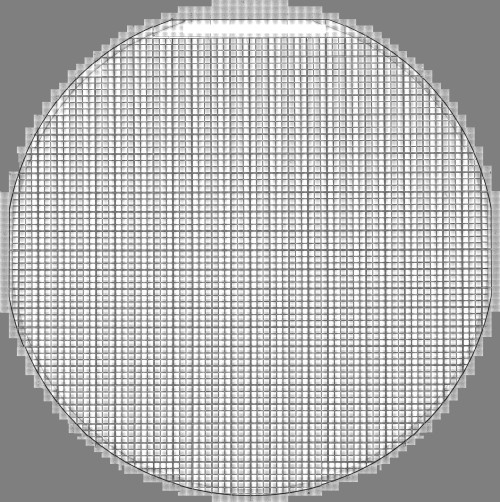}}
    \quad
	\subfloat{\includegraphics[width=0.45\linewidth]{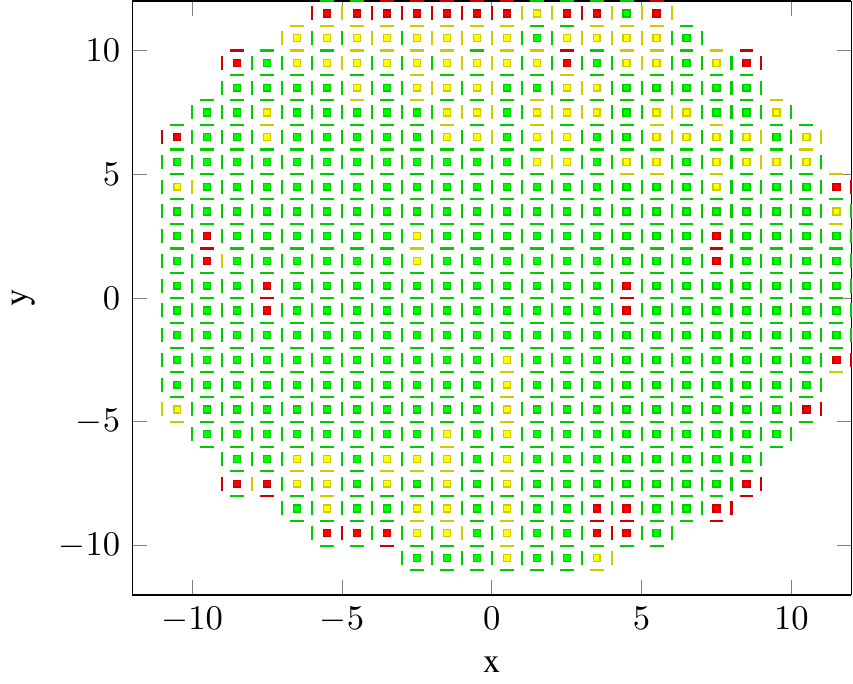}}

	\caption{Wafer overview, and chip plus street ground truth of wafers 2 - 4. The latter is visualized for good (\textcolor{green}{\textbullet}), anomaly (\textcolor{yellow}{\textbullet}) and faulty (\textcolor{red}{\textbullet}) streets. The notation of the figure is identically to Fig. \ref{fig:waferOverview}. The wafer images were again scaled down to $500\times500$ pixels to protect the intellectual property.}
	\label{fig:waferOverviewAll1}
\end{figure*}

\begin{figure*}[p]
	\centering

    \subfloat{\includegraphics[width=0.35\linewidth]{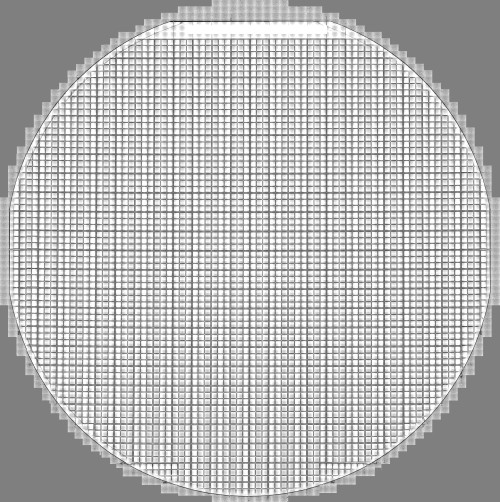}}
    \quad
	\subfloat{\includegraphics[width=0.45\linewidth]{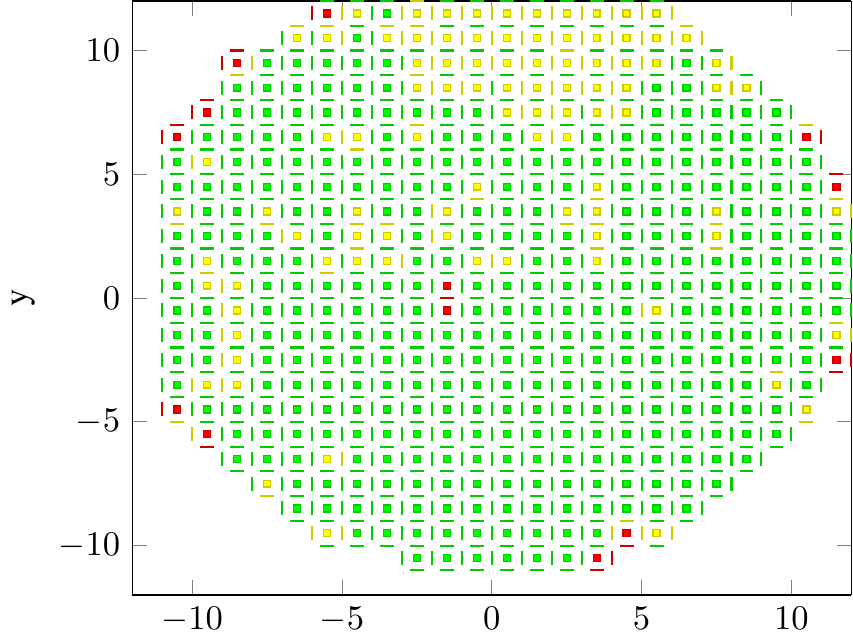}}

    \subfloat{\includegraphics[width=0.35\linewidth]{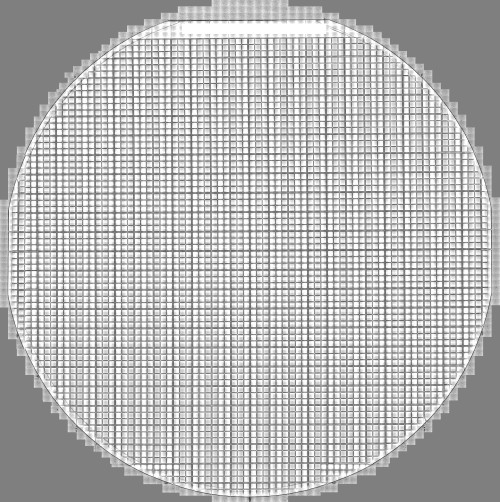}}
    \quad
	\subfloat{\includegraphics[width=0.45\linewidth]{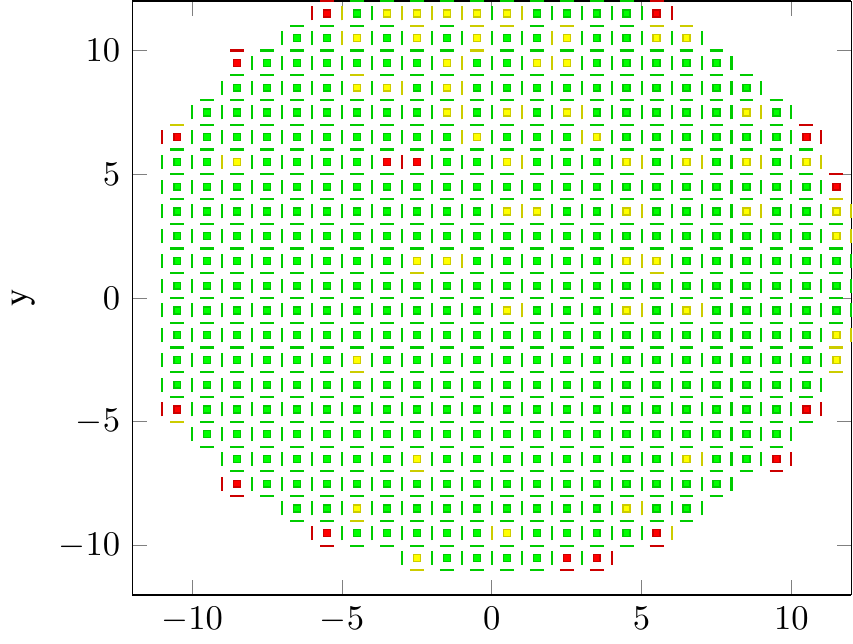}}

    \subfloat{\includegraphics[width=0.35\linewidth]{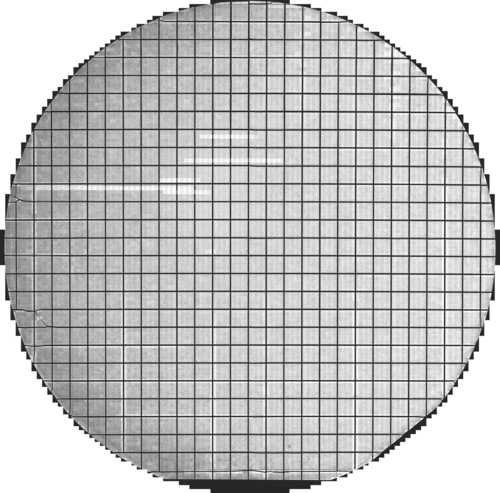}}
    \quad
	\subfloat{\includegraphics[width=0.45\linewidth]{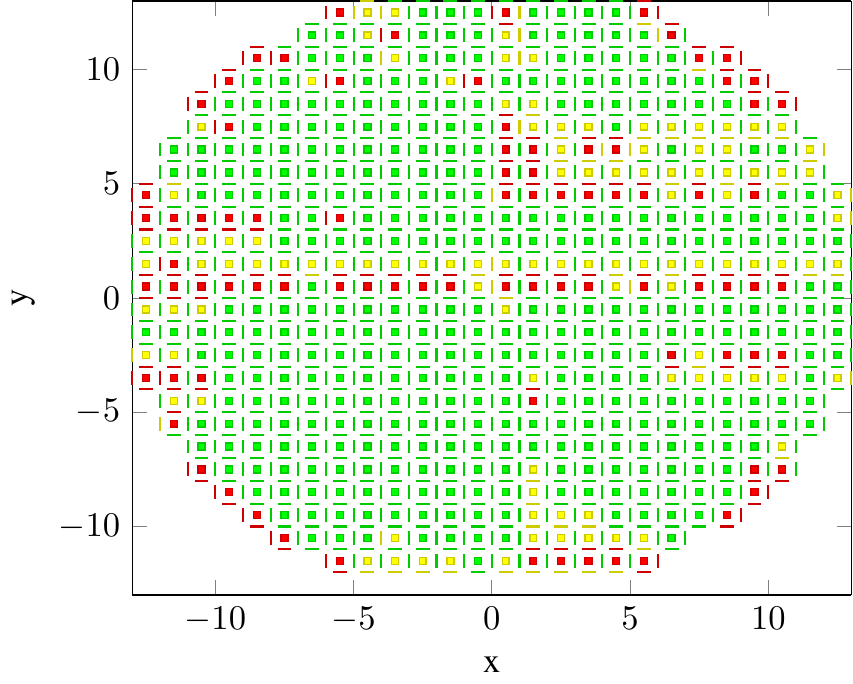}}

    \caption{Wafer overview and street ground truth of wafers 5 - 7. The figure is identically labeled to Fig. \ref{fig:waferOverviewAll1}.}
	\label{fig:waferOverviewAll3}
\end{figure*}

\begin{figure*}[p]
	\centering

    \subfloat{\includegraphics[width=0.35\linewidth]{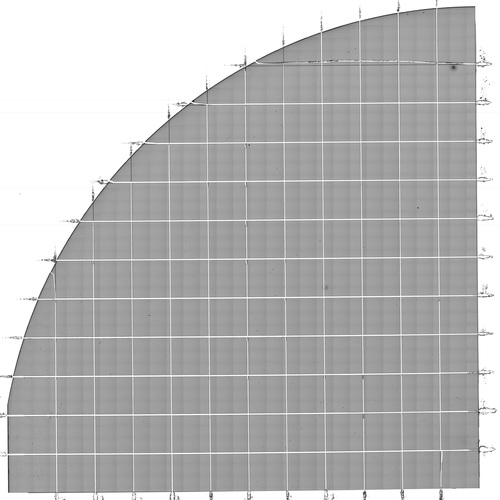}}
    \quad
	\subfloat{\includegraphics[width=0.45\linewidth]{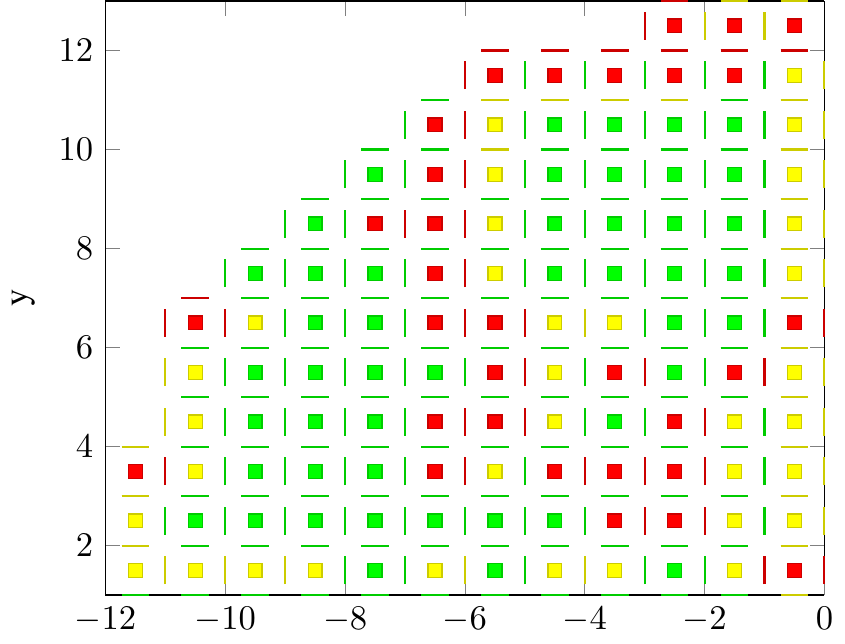}}

    \subfloat{\includegraphics[width=0.35\linewidth]{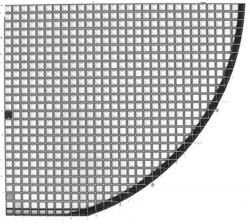}}
    \quad
	\subfloat{\includegraphics[width=0.45\linewidth]{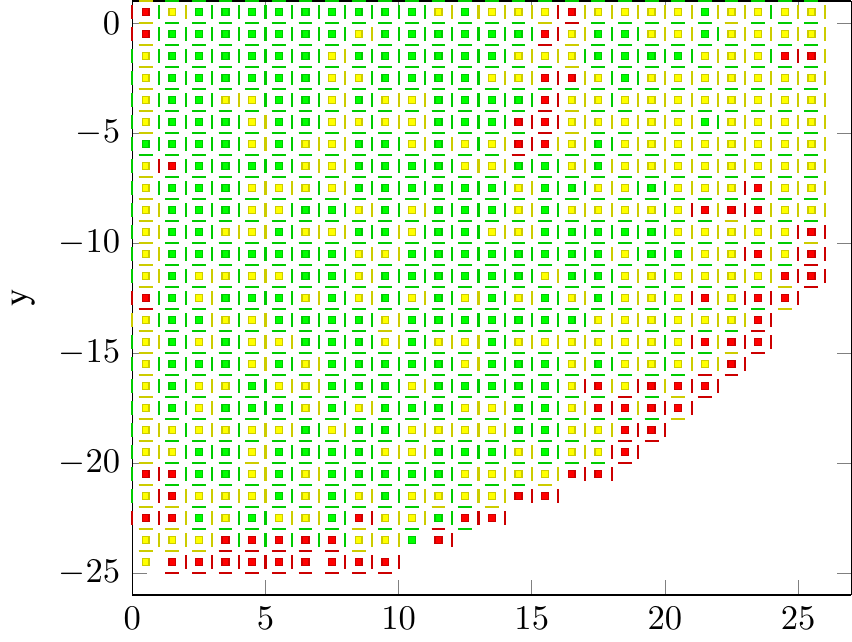}}

    \subfloat{\includegraphics[width=0.35\linewidth]{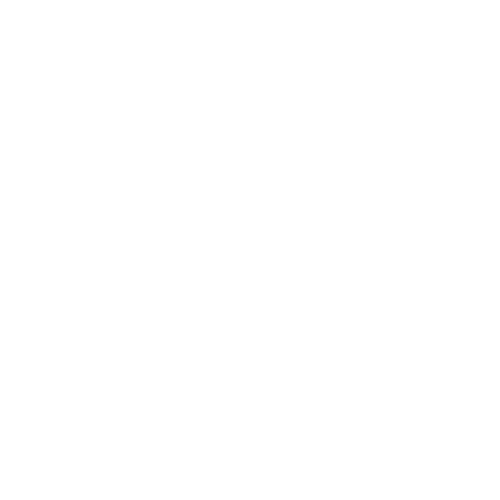}}
    \quad
	\subfloat{\includegraphics[width=0.45\linewidth]{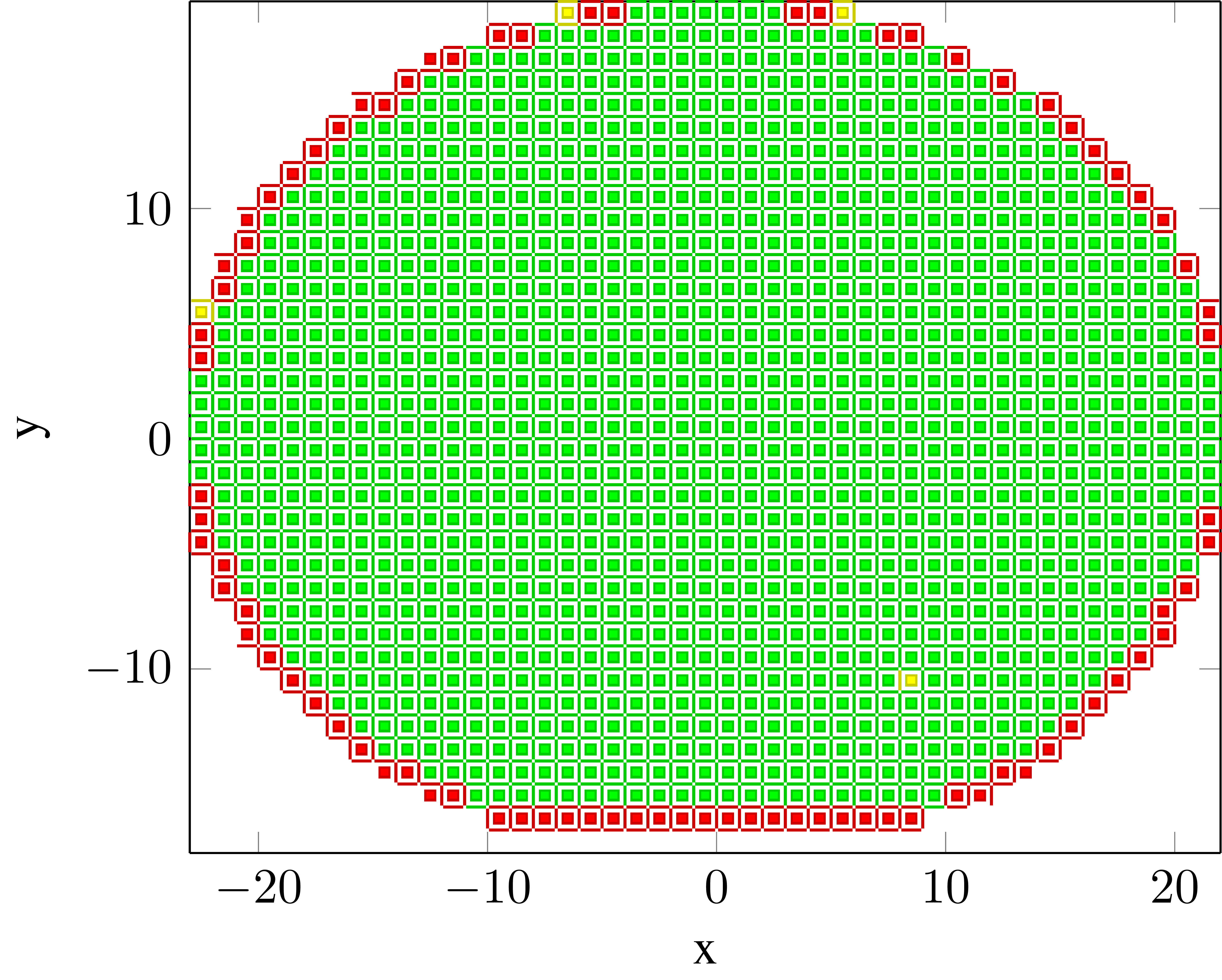}}

    \caption{Wafer overview and street ground truth of wafers 8 - 10. The figure is identically labeled to Fig. \ref{fig:waferOverviewAll1}. The wafer 10 is not available as a whole image, hence the wafer is not displayed here. We have only been provided single chip images.}
	\label{fig:waferOverviewAll5}
\end{figure*}